%% file: egpaper_for_review.tex
\PassOptionsToPackage{warn}{textcomp}
\documentclass[10pt,twocolumn,letterpaper]{article}

\usepackage{iccv}
\usepackage{times}
\usepackage{epsfig}
\usepackage{graphicx}
\usepackage{amsmath}
\usepackage{amssymb}
\usepackage{xcolor}
\usepackage{bm}
\usepackage{verbatim}
\usepackage{booktabs}
\usepackage{textcomp}
\usepackage[symbol]{footmisc}

\renewcommand{\thefootnote}{\fnsymbol{footnote}}
\usepackage[pagebackref=true,breaklinks=true,letterpaper=true,colorlinks,bookmarks=false]{hyperref}

\newcommand{\jv}[1]{{\color{blue}{\bf\sf [JV: #1]}}}

\newcommand{\mj}[1]{{\color{purple}{\bf\xf [MJ: #1]}}}

\newcommand{\name}{FastNeRF}

\iccvfinalcopy % *** Uncomment this line for the final submission

\def\iccvPaperID{7615} % *** Enter the ICCV Paper ID here

\usepackage{capt-of,etoolbox}
\makeatletter
\apptocmd\@maketitle{{\teaserfigure{}\par}}{}{}
\makeatother

\newcommand{\teaserfigure}{
    \centering
    \vspace{-2.5em}
    \includegraphics[width=\linewidth]{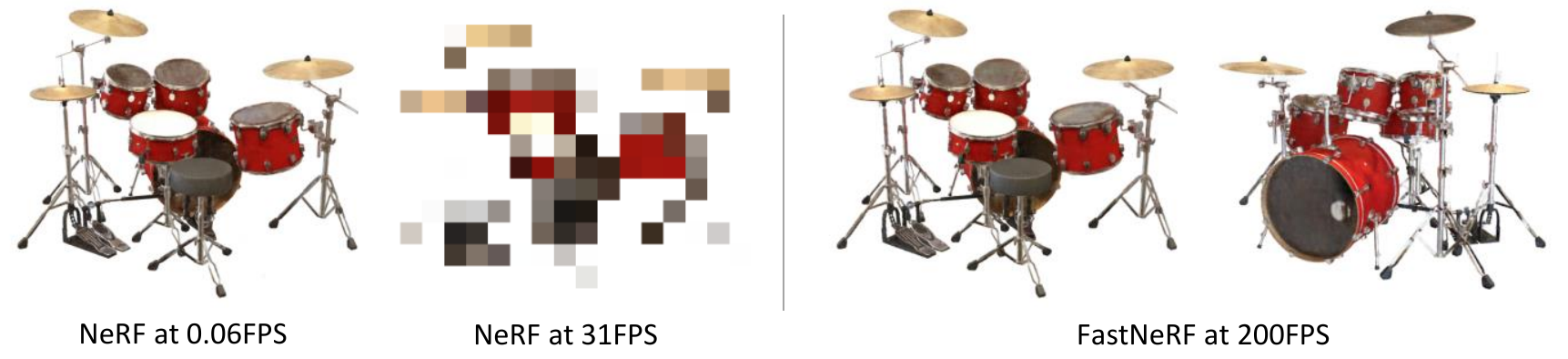}
    \captionof{figure}{\textbf{\name{}} renders high-resolution photorealistic novel views of objects at hundreds of frames per second. Comparable existing methods, such as NeRF, are orders of magnitude slower and can only render very low resolution images at interactive rates.}
    \label{fig:teaser}
    \vspace{0.2in}
}

\newcommand\blfootnote[1]{%
  \begingroup
  \renewcommand\thefootnote{}\footnote{#1}%
  \addtocounter{footnote}{-1}%
  \endgroup
}

\begin{document}

%%%%%%%%% TITLE
\title{\name: High-Fidelity Neural Rendering at 200FPS}

\author{Stephan J. Garbin\thanks{Denotes equal contribution} \and
Marek Kowalski\footnotemark[1] \and
Matthew Johnson \and
Jamie Shotton \and
Julien Valentin}

\maketitle
%\ificcvfinal\thispagestyle{empty}\fi

%%%%%%%%% ABSTRACT
\input{abstract}

%%%%%%%%% INTRODUCTION
\input{introduction}

%%%%%%%%% RELATED WORK
\input{related_work}

%%%%%%%%% METHOD
\input{method}

%%%%%%%%% IMPLEMENTATION
\input{implementation}

%%%%%%%%% EXPERIMENTS
\input{experiments}

%%%%%%%%% APPLICATIONS
\input{applications}

%%%%%%%%% CONCLUSION
\input{conclusion}

{\small
\bibliographystyle{ieee_fullname}
\bibliography{egbib}
}
\clearpage
%%%%%%%%% SUPPLEMENTARY
\input{./supplementary/supplementary}

\end{document}

%% file: abstract.tex
%!TEX root = egpaper_for_review.tex

\begin{abstract}
Recent work on Neural Radiance Fields (NeRF) showed how neural networks can be used to encode complex 3D environments that can be rendered photorealistically from novel viewpoints. Rendering these images is very computationally demanding and recent improvements are still a long way from enabling interactive rates, even on high-end hardware. Motivated by scenarios on mobile and mixed reality devices, we propose \name{}, the first NeRF-based system capable of rendering high fidelity photorealistic images at 200Hz on a high-end consumer GPU. The core of our method is a graphics-inspired factorization that allows for (i) compactly caching a deep radiance map at each position in space, (ii) efficiently querying that map using ray directions to estimate the pixel values in the rendered image. Extensive experiments show that the proposed method is 3000 times faster than the original NeRF algorithm and at least an order of magnitude faster than existing work on accelerating NeRF, while maintaining visual quality and extensibility.
\end{abstract}

%% file: introduction.tex
%!TEX root = egpaper_for_review.tex

\section{Introduction}
\blfootnote{\\ Project website: \url{https://microsoft.github.io/FastNeRF}}
Rendering scenes in real-time at photorealistic quality has long been a goal of computer graphics. Traditional approaches such as rasterization and ray-tracing
 often require significant manual effort in designing or pre-processing the scene in order to achieve both quality and speed. Recently, neural rendering \cite{styleganPaper, deepshading, codecAvatarsPaper, NeRF, lombardi2021mixture} has offered a disruptive alternative: involve a neural network
%as a critical component
in the rendering pipeline to output either images directly \cite{localLightfieldFusionPaper, NeuralVolumes, deepshading} or to model implicit functions that represent a scene appropriately \cite{implicitFunctionsShapeModelling, occupancyNetworks, deepSDF, NeRF, tancik2020fourier}. Beyond rendering, some of these approaches implicitly reconstruct a scene from static or moving cameras \cite{NeRF, stableViewSynthesis, xfields}, thereby greatly simplifying the traditional reconstruction pipelines used in computer vision.

One of the most prominent recent advances in neural rendering is Neural Radiance Fields (NeRF) \cite{NeRF} which, given a handful of images of a static scene, learns an implicit volumetric representation of the scene that can be rendered from novel viewpoints. The rendered images are of high quality and correctly retain thin structures, view-dependent effects, and partially-transparent surfaces. NeRF has inspired significant follow-up work that has addressed some of its limitations, notably extensions to dynamic scenes \cite{nerfies, gafni2020dynamic, spaceTimeNeRF}, relighting \cite{neuralReflectanceFields, deepReflectanceVolumes, srinivasan2020nerv}, and incorporation of uncertainty \cite{nerfInTheWild}.

One common challenge to all of the NeRF-based approaches is their high computational requirements for rendering images. The core of this challenge resides in NeRF's volumetric scene representation. More than 100 neural network calls are required to render a single image pixel, which translates into several seconds being required to render low-resolution images on high-end GPUs.
%Motivated by the promise of enabling new disruptive applications,
Recent explorations \cite{rebain2020derf, lindell2020autoint, neff2021donerf, liu2021neural} conducted with the aim of improving NeRF's computational requirements reduced the render time by up to 50$\times$.
While impressive, these advances are still a long way from enabling real-time rendering on consumer-grade hardware. Our work bridges this gap while maintaining quality, thereby opening up a wide range of new applications for neural rendering.  Furthermore, our method could form the fundamental building block for neural rendering at high resolutions.

To achieve this goal, we use caching to trade memory for computational efficiency. As NeRF is fundamentally a function of positions $\mathbf{p} \in \mathbb{R}^3$ and ray directions $\mathbf{d} \in \mathbb{R}^2$ to color $\bm{c} \in \mathbb{R}^3$ (RGB) and a scalar density $\sigma$, a na\"{i}ve approach would be to build a cached representation of this function in the space of its input parameters. Since $\sigma$ only depends on $\mathbf{p}$, it can be cached using existing methodologies. The color $\bm{c}$, however, is a function of both ray direction $\mathbf{d}$ \textit{and} position $\mathbf{p}$. If this 5 dimensional space were to be discretized with 512 bins per dimension, a cache of around 176 terabytes of memory would be required -- dramatically more than is available on current consumer hardware.

Ideally, we would treat directions and positions separately and thus avoid the polynomial explosion in required cache space. Fortunately this problem is not unique to NeRF; the rendering equation (modulo the wavelength of light and time) is also a function of $\mathbb{R}^5$ and solving it efficiently is the primary challenge of real-time computer graphics. As such there is a large body of research which investigates ways of approximating this function as well as efficient means of evaluating the integral. One of the fastest approximations of the rendering equation involves the use of spherical harmonics such that the integral results in a dot product of the harmonic coefficients of the material model and lighting model. Inspired by this efficient approximation, we factorize the problem into two independent functions whose outputs are combined using their inner product to produce RGB values. The first function takes the form of a Multi-Layer Perceptron (MLP) that is conditioned on position in space and returns a compact deep radiance map parameterized by $D$ components. The second function is an MLP conditioned on ray direction that produces $D$ weights for the $D$ deep radiance map components.

This factorized architecture, which we call \name{}, allows for independently caching the position-dependent and ray direction-dependent outputs.
Assuming that $k$ and $l$ denote the number of bins for positions and ray directions respectively, caching NeRF would have a memory complexity of $\mathcal{O}(k^3l^2)$. In contrast, caching \name{} would have a complexity of $\mathcal{O}(k^3*(1 + D*3) + l^2*D)$. As a result of this reduced memory complexity, \name{} can be cached in the memory of a high-end consumer GPU, thus enabling very fast function lookup times that in turn lead to a dramatic increase in test-time performance.

While caching does consume a significant amount of memory, it is worth noting that current implementations of NeRF also have large memory requirements. A single forward pass of NeRF requires performing hundreds of forward passes through an eight layer $256$ hidden unit MLP \textit{per pixel}. If pixels are processed in parallel for efficiency this consumes large amounts of memory, even at moderate resolutions. Since many natural scenes (\eg a living room, a garden) are sparse, we are able to store our cache sparsely. In some cases this can make our method actually more memory efficient than NeRF.

In summary, our main contributions are:
\begin{itemize}
	\item The first NeRF-based system capable of rendering photorealistic novel views at $200$FPS, thousands of times faster than NeRF.
	\item A graphics-inspired factorization that can be compactly cached and subsequently queried to compute the pixel values in the rendered image.
	\item A blueprint detailing how the proposed factorization can efficiently run on the GPU.
	%MK: commented-out for text compression
	%\item Extensive experiments that demonstrate the new-found speed can be achieved while maintaining quality.
\end{itemize}

%% file: related_work.tex
%!TEX root = egpaper_for_review.tex

\section{Related work}
\name{} belongs to the family of Neural Radiance Fields methods \cite{NeRF} and is trained to learn an implicit, compressed deep radiance map parameterized by position and view direction that provides color and density estimates. Our method differs from \cite{NeRF} in the structure of the implicit model, changing it in such a way that positional and directional components can be discretized and stored in sparse 3D grids.

This also differentiates our method from models that use a discretized grid at training time, such as Neural Volumes \cite{NeuralVolumes} or Deep Reflectance Volumes \cite{deepReflectanceVolumes}. Due to the memory requirements associated with generating and holding 3D volumes at training time, the output image resolution of \cite{NeuralVolumes,deepReflectanceVolumes} is limited by the maximum volume size of $128^3$. In contrast, our method uses volumes as large as $1024^3$.

Subsequent to \cite{NeuralVolumes, deepReflectanceVolumes}, the problem of parameter estimation for an MLP with low dimensional input coordinates was addressed using Fourier feature encodings. After use in \cite{NIPS2017_3f5ee243}, this was popularized in \cite{NeRF, zhong2020reconstructing} and explored in greater detail by \cite{tancik2020fourier}.

Neural Radiance Fields (NeRF) \cite{NeRF} first showed convincing compression of a light field utilizing Fourier features. Using an entirely implicit model, NeRF is not bound to any voxelized grid, but only to a specific domain.  Despite impressive results, one disadvantage of this method is that a complex MLP has to be called for every sample along each ray, meaning hundreds of MLP invocations per image pixel.

One way to speed up NeRF's inference is to scale to multiple processors. This works particularly well because every pixel in NeRF can be computed independently. For an $800 \times 800$ pixel image, JaxNeRF \cite{jaxnerf2020github} achieves an inference speed of 20.77 seconds on an Nvidia Tesla V100 GPU, 2.65 seconds on \textit{8} V100 GPUS, and 0.35 seconds on 128 second generation Tensor Processing Units. Similarly, the volumetric integration domain can be split and separate models used for each part. This approach is taken in the Decomposed Radiance Fields method \cite{rebain2020derf}, where an integration domain is divided into Voronoi cells. This yields better test scores, and a speedup of up to $3\times$.

A different way to increase efficiency is to realize that natural scenes tend to be volumetrically sparse. Thus, efficiency can be gained by skipping empty regions of space. This amounts to importance sampling of the occupancy distribution of the integration domain. One approach is to use a voxel grid in combination with the implicit function learned by the MLP, as proposed in Neural Sparse Voxel Fields (NSVFs) \cite{liu2021neural} and Neural Geometric Level of Detail \cite{takikawa2021neural} (for SDFs only), where a dynamically constructed sparse octree is used to represent scene occupancy. As a network still has to be queried inside occupied voxels, however, NSVFs takes between 1 and 4 seconds to render an $800^2$ image, with decreases to PSNR at the lower end of those timings. Our method is orders of magnitude faster in a similar scenario. 

Another way to represent the importance distribution is via depth prediction for each pixel. This approach is taken in \cite{neff2021donerf}, which is concurrent to ours and achieves roughly 15FPS for $800^2$ images at reduced quality or roughly half that for quality comparable to NeRF.

Orthogonal to this, AutoInt \cite{lindell2020autoint} showed that a neural network can be used to approximate the integrals along each ray with far fewer samples. While significantly faster than NeRF, this still does not provide interactive frame rates.

What differentiates our method from those described above is that FastNeRF's proposed decomposition, and subsequent caching, lets us avoid calls to an MLP at inference time entirely. This makes our method faster in absolute terms even on a single machine.  

It is worth noting that our method does not address training speed, as for example \cite{tancik2020learned}.  In that case, the authors propose improving the training speed of NeRF models by finding initialisation through meta learning.

Finally, there are orthogonal neural rendering strategies capable of fast inference, such as Neural Point-Based Graphics \cite{kolos2020transpr}, Mixture of Volumetric Primitives \cite{lombardi2021mixture} or Pulsar \cite{lassner2020pulsar}, which use forms of geometric primitives and rasterization. In this work, we only deal with NeRF-like implicit functional representations paired with ray tracing.

%% file: method.tex
%!TEX root = egpaper_for_review.tex

\section{Method}
\begin{figure*}[!ht]
    \centering
     \includegraphics[width=\textwidth]{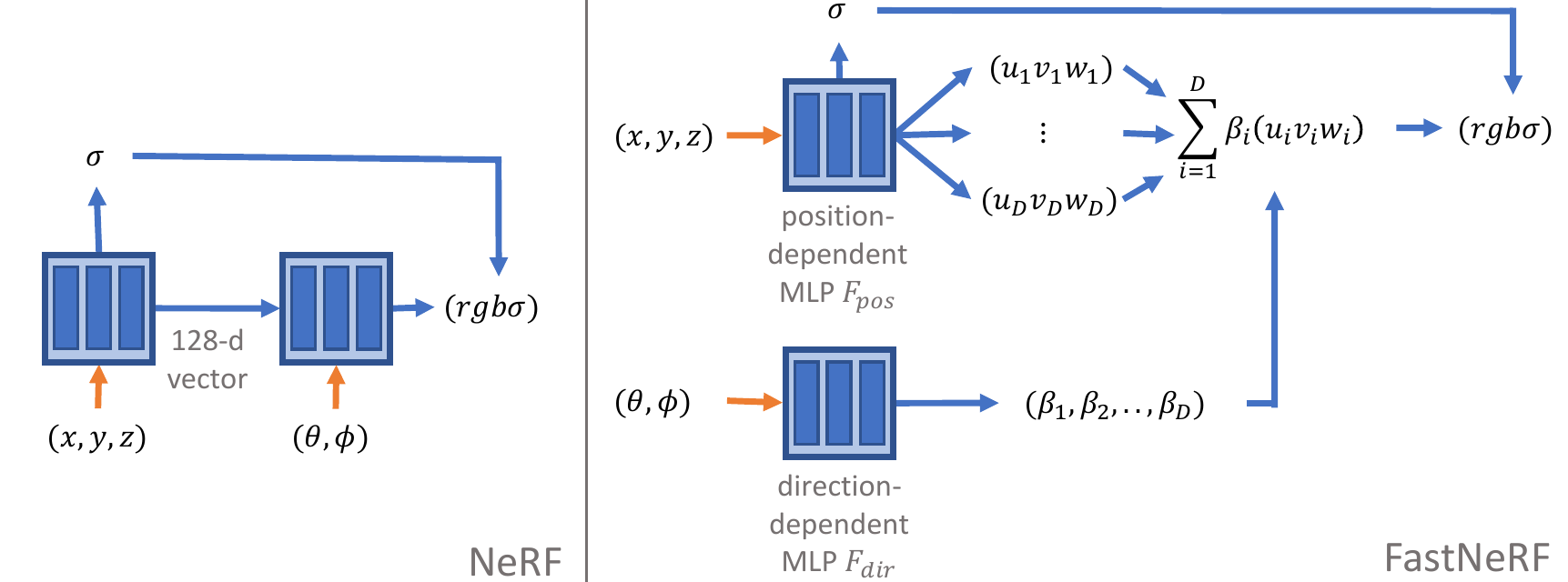}
     \caption{Left: NeRF neural network architecture. $(x,y,z)$ denotes the input sample position, $(\theta, \phi)$ denotes the ray direction and $(r,g,b,\sigma)$ are the output color and transparency values. Right: our \name{} architecture splits the same task into two neural networks that are amenable to caching. The position-dependent network $\mathcal{F}_{pos}$ outputs a deep radiance map $(\bm{u}, \bm{v}, \bm{w})$ consisting of $D$ components, while the $\mathcal{F}_{dir}$ outputs the weights for those components $(\beta_1,\dots,\beta_{D})$ given a ray direction as input.}
     \label{fig:outline}
\end{figure*}

In this section we describe \name{}, a method that is 3000 times faster than the original Neural Radiance Fields (NeRF) system \cite{NeRF} (Section \ref{sec:nerf}). This breakthrough allows for rendering high-resolution photorealistic images at over 200Hz on high-end consumer hardware. The core insight of our approach (Section \ref{sec:factorized}) consists of factorizing NeRF into two neural networks: a position-dependent network that produces a deep radiance map and a direction-dependent network that produces weights. The inner product of the weights and the deep radiance map estimates the color in the scene at the specified position and as seen from the specified direction. This architecture, which we call \name{}, can be efficiently cached (Section \ref{sec:caching}), significantly improving test time efficiency whilst preserving the visual quality of NeRF. See Figure \ref{fig:outline} for a comparison of the NeRF and \name{} network architectures.

\subsection{Neural Radiance Fields}
\label{sec:nerf}
A Neural Radiance Field (NeRF) \cite{NeRF} captures a volumetric 3D representation of a scene within the weights of a neural network. NeRF's neural network $\mathcal{F}_{NeRF}: (\bm{p}, \bm{d}) \mapsto (\bm{c}, \sigma)$ maps a 3D position $\bm{p} \in \mathbb{R}^3$ and a ray direction $\bm{d} \in \mathbb{R}^2$ to a color value $\bm{c}$ and transparency $\sigma$. See the left side of Figure \ref{fig:outline} for a diagram of $\mathcal{F}_{NeRF}$.

In order to render a single image pixel, a ray is cast from the camera center, passing through that pixel and into the scene. We denote the direction of this ray as $\bm{d}$. A number of 3D positions $(\bm{p}_1,\dotsi , \bm{p}_N)$ are then sampled along the ray between its near and far bounds defined by the camera parameters. The neural network $\mathcal{F}_{NeRF}$ is evaluated at each position $\bm{p}_i$ and ray direction $\bm{d}$ to produce color $\bm{c}_i$ and transparency $\sigma_i$. These intermediate outputs are then integrated as follows to produce the final pixel color $\bm{\hat{c}}$:
\begin{equation}
\label{eq:volume_rendering}
    \bm{\hat{c}} = \sum^{N}_{i=1} T_i (1-\exp (-\sigma_i \delta_i)) \bm{c}_i,
\end{equation}
where $T_i = \exp (-\sum_{j=i}^{i-1} \sigma_j \delta_j)$ is the transmittance and $\delta_i= \left( \bm{p}_{i+1} - \bm{p}_i \right)$ is the distance between the samples. Since $\mathcal{F}_{NeRF}$ depends on ray directions, NeRF has the ability to model viewpoint-dependent effects such as specular reflections, which is one key dimension in which NeRF improves upon traditional 3D reconstruction methods.

Training a NeRF network requires a set of images of a scene as well as the extrinsic and intrinsic parameters of the cameras that captured the images. In each training iteration, a subset of pixels from the training images are chosen at random and for each pixel a 3D ray is generated. Then, a set of samples is selected along each ray and the pixel color $\bm{\hat{c}}$ is computed using Equation \eqref{eq:volume_rendering}. The training loss is the mean squared difference between $\bm{\hat{c}}$ and the ground truth pixel value. For further details please refer to \cite{NeRF}.

While NeRF renders photorealistic images, it requires calling $\mathcal{F}_{NeRF}$ a large number of times to produce a single image. With the default number of samples per pixel $N=192$ proposed in NeRF \cite{NeRF}, nearly 400 million calls to $\mathcal{F}_{NeRF}$ are required to compute a single high definition (1080p) image. Moreover the intermediate outputs of this process would take hundreds of gigabytes of memory. Even on high-end consumer GPUs, this constrains the original method to be executed over several batches even for medium resolution ($800 \times 800$) images, leading to additional computational overheads.

\subsection{Factorized Neural Radiance Fields}
\label{sec:factorized}

\iffalse
\mj{This is a rewording of the section in the Introduction. Either that section needs to be pared down and use a forward reference, or it should be expanded and this should utilize a backwards reference. At the moment it is a needless duplication.}
\jv{the content is ~15lines in each case which present the intuition at a high level in the intro and at a lower level here. In my opinion this is fine - commenting for now but feel free to push back}
\mj{I defer to your judgement :) }
\fi

Taking a step away from neural rendering for a moment, we recall that in traditional computer graphics, the rendering equation \cite{pathTracingPaper} is an integral of the form
\begin{equation}
    L_o(\bm{p}, \bm{d}) = \int_{\Omega} f_r(\bm{p},\bm{d}, \bm{\omega_i})L_i(\bm{p},\bm{\omega_i})(\bm{\omega_i} \cdot \bm{n})d\bm{\omega_i},
\end{equation}
where $L_o(\bm{p}, \bm{d})$ is the radiance leaving the point $\bm{p}$ in direction $\bm{d}$, $f_r(\bm{p},\bm{d},\bm{\omega_i})$ is the reflectance function capturing the material properties at position $\bm{p}$, $L_i(\bm{p},\bm{\omega_i})$ describes the amount of light reaching $\bm{p}$ from direction $\bm{\omega_i}$, and $\bm{n}$ corresponds to the direction of the surface normal at $\bm{p}$. Given its practical importance, evaluating this integral efficiently has been a subject of active research for over three decades \cite{pathTracingPaper, misVeachPaper, VCMPaper, pharr2016physically}. One efficient way to evaluate the rendering equation is to approximate $f_r(\bm{p},\bm{d},\bm{\omega_i})$ and $L_i(\bm{p},\bm{\omega_i})$ using spherical harmonics \cite{originalSphericalHarmonics, sphericalHarmonicsRecent}. In this case, evaluating the integral boils down to a dot product between the coefficients of both spherical harmonics approximations. 

With this insight in mind, we can return to neural rendering. In the case of NeRF, $\mathcal{F}_{NeRF}: (\bm{p}, \bm{d})$ can also be interpreted as returning the radiance leaving point $\bm{p}$ in direction $\bm{d}$. This key insight leads us to propose a new network architecture for NeRF, which we call \name{}. This novel architecture consists in splitting NeRF's neural network $\mathcal{F}_{NeRF}$ into two networks: one that depends only on positions $\bm{p}$ and one that depends only on the ray directions $\bm{d}$. Similarly to evaluating the rendering equation using spherical harmonics, our position-dependent $\mathcal{F}_{pos}$ and direction-dependent $\mathcal{F}_{dir}$ functions produce outputs that are combined using the dot product to obtain the color value at position $\bm{p}$ observed from direction $\bm{d}$. Crucially, this factorization splits a single function that takes inputs in $\mathbb{R}^5$ into two functions that take inputs in $\mathbb{R}^3$ and $\mathbb{R}^2$. As explained in the following section, this makes caching the outputs of the network possible and allows accelerating NeRF by 3 orders of magnitude on consumer hardware. See Figure \ref{fig:bar_chart} for a visual representation of the achieved speedup.

The position-dependent and direction-dependent functions of \name{} are defined as follows:
\begin{equation}
    \mathcal{F}_\mathrm{pos}: \bm{p} \mapsto \{ \sigma, (\bm{u}, \bm{v}, \bm{w}) \},
\end{equation}
\begin{equation}
    \mathcal{F}_\mathrm{dir}: \bm{d} \mapsto \bm{\beta},
\end{equation}
where $\bm{u}, \bm{v}, \bm{w}$ are $D$-dimensional vectors that form a deep radiance map describing the view-dependent radiance at position $\bm{p}$. The output of $\mathcal{F}_{dir}$, $\bm{\beta}$, is a $D$-dimensional vector of weights for the $D$ components of the deep radiance map. The inner product of the weights and the deep radiance map
\begin{equation}
\label{eq:inner_product}
    \bm{c} = (r,g,b) = \sum^{D}_{i=1} \beta_i (u_i,v_i,w_i) = \bm{\beta}^T \cdot (\bm{u}, \bm{v}, \bm{w}) 
\end{equation}
results in the estimated color $\bm{c}=(r,g,b)$ at position $\bm{p}$ observed from direction $\bm{d}$. 

When the two functions are combined using Equation \eqref{eq:inner_product}, the resulting function $\mathcal{F}_\mathrm{FastNeRF} (\bm{p}, \bm{d}) \mapsto (\bm{c}, \sigma)$ has the same signature as the function $\mathcal{F}_{NeRF}$ used in NeRF. This effectively allows for using the \name{} architecture as a drop-in replacement for NeRF's architecture at runtime.

\subsection{Caching}
\label{sec:caching}
Given the large number of samples that need to be evaluated to render a single pixel, the cost of computing $\mathcal{F}$ dominates the total cost of NeRF rendering. Thus, to accelerate NeRF, one could attempt to reduce the test-time cost of $\mathcal{F}$ by caching its outputs for a set of inputs that cover the space of the scene. The cache can then be evaluated at a fraction of the time it takes to compute $\mathcal{F}$.

For a trained NeRF model, we can define a bounding box $\mathcal{V}$ that covers the entire scene captured by NeRF. We can then uniformly sample $k$ values for each of the 3 world-space coordinates $(x,y,z) = \bm{p}$ within the bounds of $\mathcal{V}$. Similarly, we uniformly sample $l$ values for each of the ray direction coordinates $(\theta, \phi) = \bm{d}$ with $\theta \in \langle 0, \pi \rangle$ and $\phi \in \langle 0, 2\pi \rangle$. The cache is then generated by computing $\mathcal{F}$ for each combination of sampled $\bm{p}$ and $\bm{d}$.

The size of such a cache for a \textit{standard} NeRF model with $k=l=1024$ and densely stored 16-bit floating point values is approximately 5600 Terabytes. Even for highly sparse volumes, where one would only need to keep ~1\% of the outputs, the size of the cache would still severely exceed the memory capacity of consumer-grade hardware. This huge memory requirement is caused by the usage of both $\bm{d}$ and $\bm{p}$ as input to $\mathcal{F}_{NeRF}$. A separate output needs to be saved for each combination of $\bm{d}$ and $\bm{p}$, resulting in a memory complexity in the order of $\mathcal{O}(k^3l^2)$. 

Our \name{} architecture makes caching feasible. For $k=l=1024,D=8$ the size of two dense caches holding $\{ \sigma, (\bm{u}, \bm{v}, \bm{w}))\}$ and $\bm{\beta}$ would be approximately 54 GB. For moderately sparse volumes, where 30\% of space is occupied, the memory requirement is low enough to fit into either the CPU or GPU memory of consumer-grade machines. In practice, the choice $k$ and $l$ depends on the scene size and the expected image resolution. For many scenarios a smaller cache of $k=512, l=256$ is sufficient, lowering the memory requirements further. Please see the supplementary materials for formulas used to calculate cache sizes for both network architectures.  

\begin{comment}
The amount of memory $M_{NeRF},M_{factor}$ required to store such a cache for standard and factorized NeRF models respectively can be computed as follows:
\begin{equation}
    M_{NeRF} = \alpha(s_{\sigma} + s_{rgb})k^3l^2,
\end{equation}
where $s_{\sigma}, s_{rgb}$ are the sizes of the stored transparency and RGB values in bits and $\alpha \in \langle 0, 1)$ is the inverse volume sparsity, where $\alpha=1$ would indicate a dense volume. 
The same quantity can be computed for a Factorized NeRF model as follows:
\begin{equation}
     M_{factor} = \alpha \left( (D \cdot s_{abc} + s_{\sigma})k^3 \right) + D \cdot s_{\beta} l^2,
\end{equation}
where $s_{abc}, s_{\beta}$ are the sizes of the stored $(a_i,b_i,c_i)$ values and the weights $\beta_i$.

For $k=l=1024$, $(r,g,b)$ stored as individual bytes and all other values stored as half-precision floats  the cache size would be
\begin{equation}
    M_{NeRF} = \alpha \cdot 5,629,499,534,213,120 \approx \alpha \cdot 5.6PB,
\end{equation} 
for NeRF and
\begin{equation}
    M_{factor}=\alpha \cdot 105,226,698,752 + 33,554,532 \approx \alpha \cdot 105GB,
\end{equation}
for Factorized NeRF.

Even for highly sparse volumes, where $\alpha=0.01$, the cache would take nearly 60TB, which exceeds the memory capacity of consumer-grade hardware. The high memory requirement makes this approach impractical when used with the standard NeRF architecture.
\end{comment}

%% file: implementation.tex
%!TEX root = egpaper_for_review.tex

\section{Implementation}
\begin{figure}
    \centering
     \includegraphics[width=\linewidth]{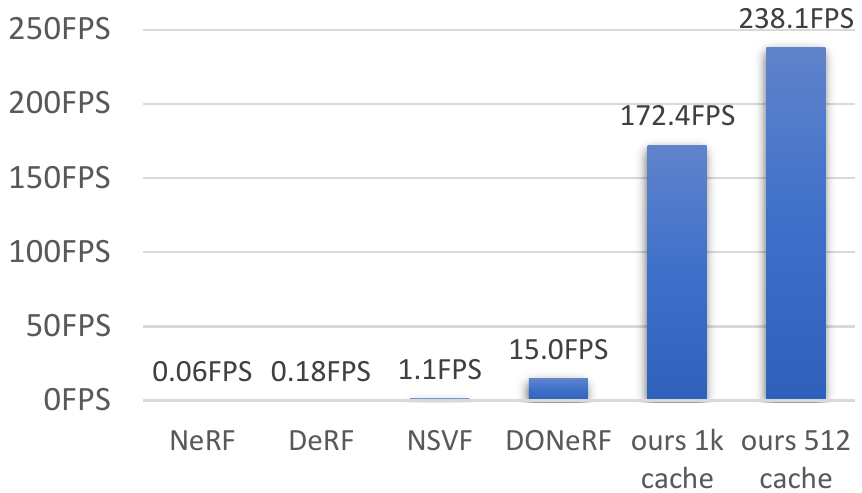}
     \caption{Speed evaluation of our method and prior work \cite{NeRF,rebain2020derf,liu2021neural,neff2021donerf} on the Lego scene from the Realistic 360 Synthetic \cite{NeRF} dataset, rendered at $800\times800$ pixels. For previous methods, when numbers for the Lego scene were not available, we used an optimistic approximation.}
     \label{fig:bar_chart}
\end{figure}

\label{sec:implementation}
\begin{figure*}
    \centering
     \includegraphics[width=\textwidth]{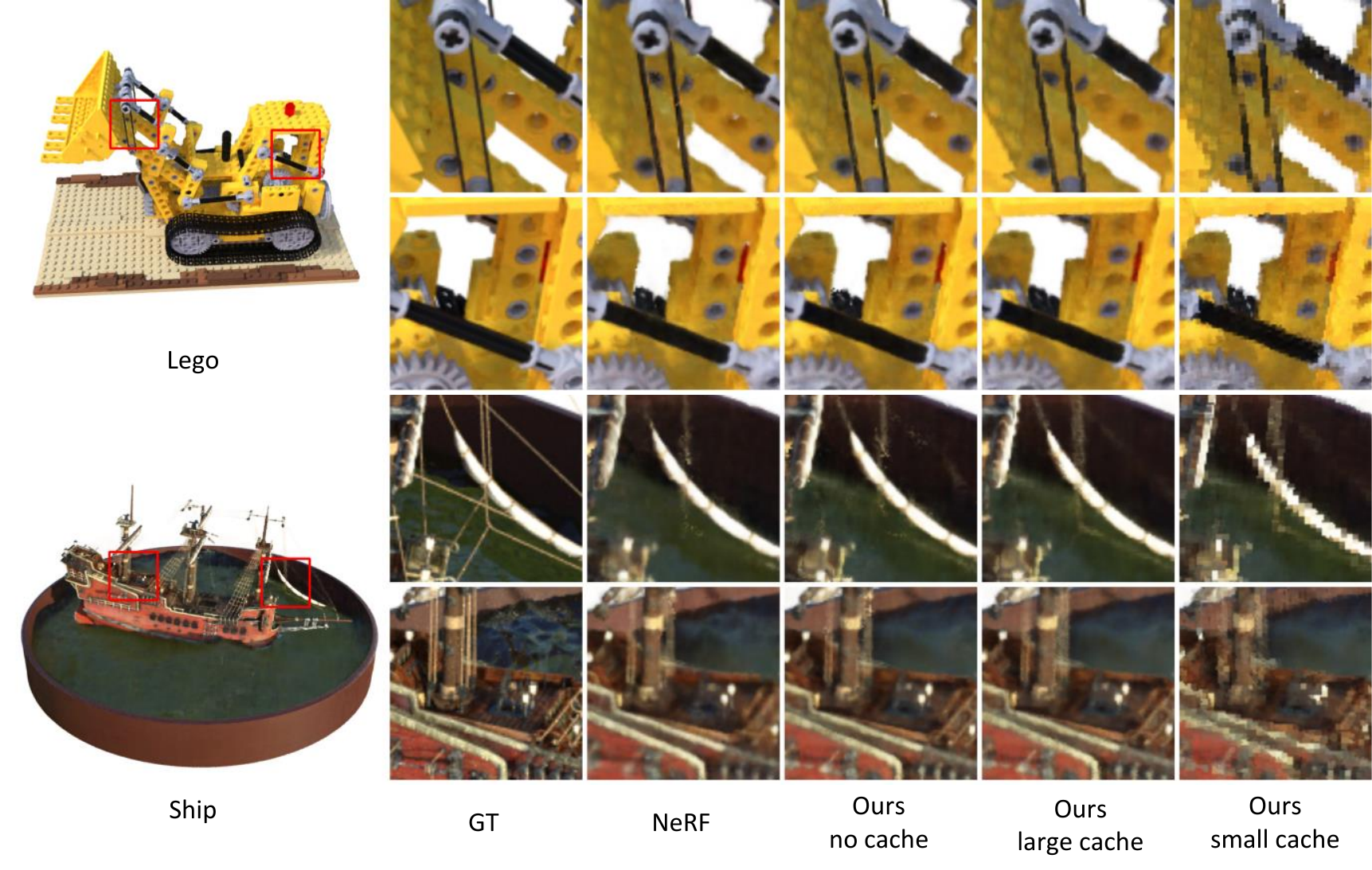}
     \caption{Qualitative comparison of our method vs NeRF on the dataset of \cite{NeRF} at $800^2$ pixels using 8 components. \textit{Small cache} refers to our method cached at $256^3$, and \textit{large cache} at $768^3$. Varying the cache size allows for trading compute and memory for image quality resembling levels of detail (LOD) in traditional computer graphics.}
     \label{fig:qualitativeA}
\end{figure*}

\begin{figure*}
    \centering
     \includegraphics[width=\textwidth]{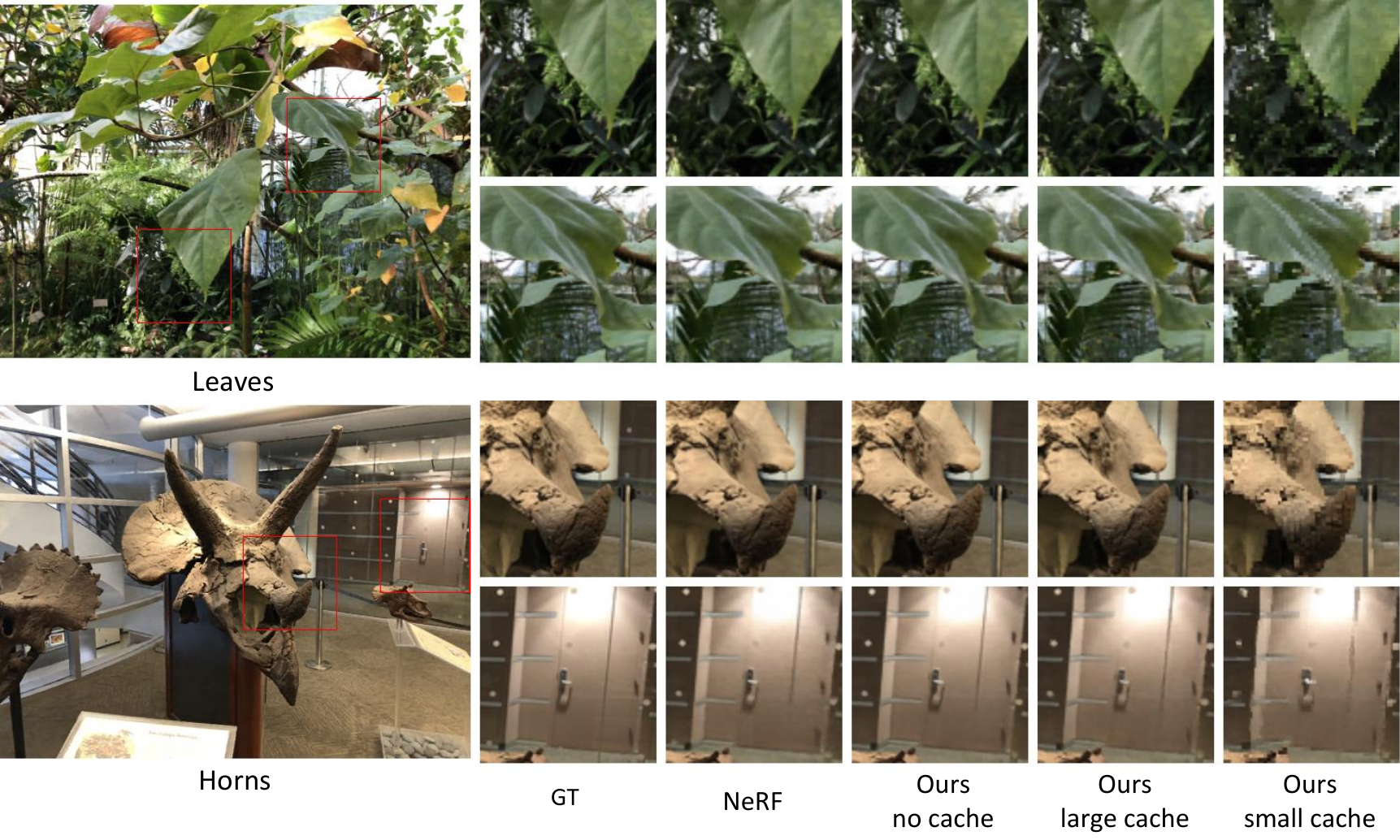}
     \caption{Qualitative comparison of our method vs NeRF on the dataset of \cite{localLightfieldFusionPaper} at $504\times378$ pixels using 6 factors. \textit{Small cache} refers to our method cached at $256^3$, and \textit{large cache} at $768^3$.}
     \label{fig:qualB}
\end{figure*}

Aside from using the proposed \name{} architecture described in Section \ref{sec:factorized}, training \name{} is identical to training NeRF \cite{NeRF}. We model $\mathcal{F}_\mathrm{pos}$ and $\mathcal{F}_\mathrm{view}$ of \name{} using 8 and 4 layer MLPs respectively, with positional encoding applied to inputs \cite{NeRF}. The 8-layer MLP has $384$ hidden units, and we use $128$ for the view-predicting 4-layer network. The only exception to this is the \textit{ficus} scene, where we found a $256$ MLP gave better results for the 8-layer MLP. While this makes the network larger than the baseline, cache lookup speed is not influenced by MLP complexity. Similarly to NeRF, we parameterize the view direction as a 3-dimensional unit vector. Other training-time features and settings including the coarse/fine networks and sample count $N$ follow the original NeRF work \cite{NeRF} (please see supplementary materials).

At test time, both our method and NeRF take a set of camera parameters as input which are used to generate a ray for each pixel in the output. A number of samples are then produced along each ray and integrated following Section \ref{sec:nerf}. While \name{} can be executed using its neural network representation, its performance is greatly improved when cached. To further improve performance, we use hardware accelerated ray tracing \cite{parker2010optix} to skip empty space, starting to integrate points along the ray only after the first hit with a collision mesh derived from the density volume, and visiting every voxel thereafter until a ray's transmittance is saturated. The collision mesh can be computed from a sign-distance function derived from the density volume using Marching Cubes \cite{marchingCubes}. For volumes greater than $512^3$, we first downsample the volume by a factor of two to reduce mesh complexity. We use the same meshing and integration parameters for all scenes and datasets. Value lookup uses nearest neighbour interpolation for $\mathcal{F}_\mathrm{pos}$ and trilinear sampling for $\mathcal{F}_\mathrm{view}$, the latter of which can be processed with a Gaussian or other kernel to `smooth' or edit directional effects. We note that the meshes and volumes derived from the cache can also be used to provide approximations to shadows and global illumination in a path-tracing framework.

%% file: experiments.tex
%!TEX root = egpaper_for_review.tex

\section{Experiments}
\begin{table*}[]
\centering
\caption{We compare NeRF to our method when not cached to a grid, and when cached at a high resolution in terms of PSNR, SSIM and LPIPS, and also provide the average speed in ms of our method when cached. LLFF denotes the dataset of \cite{localLightfieldFusionPaper} at $504\times378$ pixels, the other dataset is that of \cite{NeRF} at $800^2$ pixels. We use $8$ components and a $1024^3$ cache for the NeRF Synthetic 360 scenes, and 6 components at $768^3$ for LLFF. Please see the supplementary material for more detailed results.}

\begin{tabular}{@{}l|lll | lll | llll@{}}
\toprule
\textbf{Scene} & \multicolumn{3}{c|}{\textbf{NeRF}}              & \multicolumn{3}{c|}{\textbf{Ours - No Cache}}   & \multicolumn{3}{c}{\textbf{Ours - Cache}} & \textbf{Speed} \\

               & \textit{PSNR}$\uparrow$ & \textit{SSIM}$\uparrow$ & \textit{LPIPS}$\downarrow$ & \textit{PSNR}$\uparrow$ & \textit{SSIM}$\uparrow$ & \textit{LPIPS}$\downarrow$ & \textit{PSNR}$\uparrow$   & \textit{SSIM}$\uparrow$  & \textit{LPIPS}$\downarrow$ & \\
               \hline
\textbf{Nerf Synthetic} & $29.54$dB  & $0.94$  & $0.05$  & $29.155$dB  & $0.936$  & $0.053$  & $29.97$dB  & $0.941$  & $0.053$  & $4.2$ms \\
\textbf{LLFF} & $27.72$dB  & $0.88$  & $0.07$  & $27.958$dB  & $0.888$  & $0.063$  & $26.035$dB  & $0.856$  & $0.085$  & $1.4$ms \\
\end{tabular}
\label{tab:mainQuantitative}
\end{table*}

\begin{table*}
\centering
\caption{Speed comparison of our method vs NeRF in ms. The \textit{Chair} and \textit{Lego} scenes are at rendered at $800^2$ resolution, \textit{Horns} and \textit{Leaves} scenes at $504\times378$. Our method never drops below 100FPS when cached, and is often significantly faster. Please note that our method is slower when not cached due to using larger MLPs; it performs identically to NeRF when using $256$ hidden units. We do not compute the highest resolution cache for LLFF scenes because these are less sparse due to the use of normalized device coordinates.}
\begin{tabular}{@{}lcccccccc@{}}
\midrule
\textbf{Scene}  & NeRF & Ours - No Cache & $256^3$ & $384^3$ & $512^3$ & $768^3$ & $1024^3$ & Speedup over NeRF\\ \midrule
\textbf{Chair} & $17.5\textbf{K} $ & $28.2\textbf{K} $ & $0.8$ & $1.1$ & $1.4$ & $2.0$ & $2.7$ & $6468\times$ - $21828\times$\\ \midrule
\textbf{Lego} & $17.5\textbf{K} $ & $28.2\textbf{K} $ & $1.5$ & $2.1$ & $2.8$ & $4.2$ & $5.6$ & $3118\times$ - $11639\times$\\ \midrule
\textbf{Horns*} & $3.8\textbf{K} $ & $6.2\textbf{K} $ & $0.5$ & $0.7$ & $0.9$ & $1.2$ & - & $3183\times$ - $7640\times$\\ \midrule
\textbf{Leaves*} & $3.9\textbf{K} $ & $6.3\textbf{K} $ & $0.6$ & $0.8$ & $1.0$ & $1.5$ & - & $2626\times$ - $6566\times$\\ \midrule
\end{tabular}
\label{tab:speed}
\end{table*}

\begin{table*}[]
\centering
\caption{Influence of the number of components and grid resolution on PSNR and memory required for caching the ship scene. Note how more factors increase grid sparsity. We find that 8 or 6 components are a reasonable compromise in practice.}
\begin{tabular}{@{}l|ll|ll|ll|ll|ll@{}}
\toprule
\textbf{Factors} & \multicolumn{2}{c|}{\textbf{No Cache}} & \multicolumn{2}{c|}{\textbf{$256^3$}} & \multicolumn{2}{c|}{\textbf{$384^3$}} & \multicolumn{2}{c|}{\textbf{$512^3$}} & \multicolumn{2}{c}{\textbf{$768^3$}} \\
\textbf{}        & \textit{PSNR}$\uparrow$    & \textit{Memory}   & \textit{PSNR}$\uparrow$    & \textit{Memory}   & \textit{PSNR}$\uparrow$    & \textit{Memory}   & \textit{PSNR}$\uparrow$    & \textit{Memory}   & \textit{PSNR}$\uparrow$             & \textit{Memory}           \\ \hline
\textbf{4} & $27.11$dB  & -  & $24.81$dB  & $0.34$Gb  & $26.29$dB  & $0.61$Gb  & $26.94$dB  & $1.09$Gb  & $27.54$dB  & $2.51$Gb \\ %\hline
\textbf{6} & $27.12$dB  & -  & $24.82$dB  & $0.5$Gb  & $26.34$dB  & $0.93$Gb  & $27.0$dB  & $1.67$Gb  & $27.58$dB  & $4.1$Gb \\ %\hline
\textbf{8} & $27.24$dB  & -  & $24.89$dB  & $0.71$Gb  & $26.42$dB  & $1.41$Gb  & $27.1$dB  & $2.7$Gb  & $27.72$dB  & $7.15$Gb \\ %\hline
\textbf{16} & $27.68$dB  & -  & $25.07$dB  & $1.2$Gb  & $26.77$dB  & $2.08$Gb  & $27.55$dB  & $3.72$Gb  & $28.3$dB  & $9.16$Gb \\ %\hline
\end{tabular}
\label{tab:mainFactors}
\end{table*}

We evaluate our method quantitatively and qualitatively on the Realistic 360 Synthetic \cite{NeRF} and Local Light Field Fusion (LLFF) \cite{localLightfieldFusionPaper} (with additions from \cite{NeRF}) datasets used in the original NeRF paper. While the NeRF synthetic dataset consists of 360 degree views of complex objects, the LLFF dataset consist of forward-facing scenes, with fewer images. In all comparisons with NeRF we use the same training parameters as described in the original paper \cite{NeRF}.

To assess the rendering quality of \name{} we compare its outputs to GT using Peak Signal to Noise Ratio (PSNR), Structural Similarity (SSIM) \cite{SSIMPaper} and perceptual LPIPS \cite{LPIPSPaper}. We use the same metrics in our ablation study, where we evaluate the effects of changing the number of components $D$ and the size of the cache. All speed comparisons are run on one machine that is equipped with an Nvidia RTX 3090 GPU.

We keep the size of the view-dependent cache $\mathcal{F}_\mathrm{view}^\mathrm{cache}$ the same for all experiments at a base resolution of $384^3$. It is the base resolution of the RGBD cache $\mathcal{F}_\mathrm{pos}^\mathrm{cache}$ that represents the main trade-off in complexity of runtime and memory vs image quality. Across all results, grid resolution refers to the number of divisions of the longest side of the bounding volume surrounding the object of interest.

\textbf{Rendering quality:} Because we adopt the inputs, outputs, and training of NeRF, we retain the compatibility with ray-tracing \cite{whitted2005improved, pathTracingPaper}  and the ability to only specify loose volume bounds. At the same time, our factorization and caching do not affect the character of the rendered images to a large degree. Please see Figure~\ref{fig:qualitativeA} and Figure~\ref{fig:qualB} for qualitative comparisons, and Table~\ref{tab:mainQuantitative} for a quantitative evaluation. Note that some aliasing artefacts can appear since our model is cached to a grid \textit{after} training. This explains the decrease across all metrics as a function of the grid resolution as seen in Table~\ref{tab:mainQuantitative}. However, both NeRF and \name{} mitigate multi-view inconsistencies, ghosting and other such artifacts from multi-view reconstruction. Table~\ref{tab:mainQuantitative} demonstrates that at high enough cache resolutions, our method is capable of the same visual quality as NeRF. Figure~\ref{fig:qualitativeA} and Figure~\ref{fig:qualB} further show that smaller caches appear `pixelated' but retain the overall visual characteristics of the scene. This is similar to how different levels of detail are used in computer graphics \cite{loDBook}, and an important innovation on the path towards neurally rendered worlds.

\textbf{Cache Resolution:}
As shown in Table~\ref{tab:mainQuantitative}, our method matches or outperforms NeRF on the dataset of synthetic objects at $1024^3$ cache resolution. Because our method is fast enough to visit every voxel (as opposed to the fixed sample count used in NeRF), it can sometimes achieve better scores by not missing any detail. We observe that a cache of $512^3$ is a good trade-off between perceptual quality, memory and rendering speed for the synthetic dataset. For the LLFF dataset, we found a $768^3$ cache to work best. For the \textit{Ship} scene, an ablation over the grid size, memory, and PSNR is shown in Table~\ref{tab:mainFactors}.

For the view-filling LLFF scenes at $504\times378$ pixels, our method sees a slight decrease in metrics when cached at $768^3$, but still produces qualitatively compelling results as shown in Figure~\ref{fig:qualB}, where intricate detail is clearly preserved.

\textbf{Rendering Speed:} When using a grid resolution of $768^3$, \name{} is on average more than $3000\times$ faster than NeRF at largely the same perceptual quality. See Table~\ref{tab:speed} for a breakdown of run-times across several resolutions of the cache and Figure \ref{fig:bar_chart} for a comparison to other NeRF extensions in terms of speed. Note that we log the time it takes our method to fill an RGBA buffer of the image size with the final pixel values, whereas the baseline implementation needs to perform various steps to reshape samples back into images. Our CUDA kernels are not highly optimized, favoring flexibility over maximum performance. It is reasonable to assume that advanced code optimization and further compression of the grid values could lead to further reductions in compute time.

\textbf{Number of Components:} While we can see a theoretical improvement of roughly $0.5$dB  going from $8$ to $16$ components, we find that $6$ or $8$ components are sufficient for most scenes. As shown in Table~\ref{tab:mainFactors}, our cache can bring out fine details that are lost when only a fixed sample count is used, compensating for the difference. Table~\ref{tab:mainFactors} also shows that more components tend to be increasingly sparse when cached, which compensates somewhat for the additional memory usage. Please see the supplementary material for more detailed results on other scenes.

%% file: applications.tex
%!TEX root = egpaper_for_review.tex

\section{Application}
\begin{figure}
    \centering
     \includegraphics[width=\linewidth]{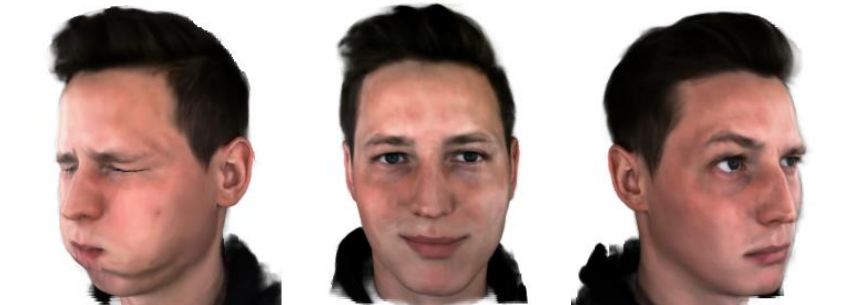}
     \caption{Face images rendered using \name{} combined with a deformation field network \cite{park2020deformable}. Thanks to the use of \name{}, expression-conditioned images can be rendered at 30FPS.}
     \label{fig:application}
\end{figure}

We demonstrate a proof-of-concept application of \name{} for a telepresence scenario by first gathering a dataset of a person performing facial expressions for about 20s in a multi-camera rig consisting of 32 calibrated cameras. We then fit a 3D face model similar to FLAME \cite{flame} to obtain expression parameters for each frame. Since the captured scene is not static, we train a \name{} model of the scene jointly with a deformation model \cite{park2020deformable} conditioned on the expression data. The deformation model takes the samples $\bm{p}$ as input and outputs their updated positions that live in a canonical frame of reference modeled by \name{}.% 

We show example outputs produced using this approach in Figure \ref{fig:application}. 
This method allows us to render $300 \times 300$ pixel images of the face at 30 fps on a single Nvidia Tesla V100 GPU -- around 50 times faster than a setup with a NeRF model. At the resolution we achieve, the face expression is clearly visible and the high frame rate allows for real-time rendering, opening the door to telepresence scenarios. The main limitation in terms of speed and image resolution is the deformation model, which needs to be executed for a large number of samples. We employ a simple pruning method, detailed in the supplementary material, to reduce the number of samples.% that have to go through the deformation model.

Finally, note that while we train this proof-of-concept method on a dataset captured using multiple cameras, the approach can be extended to use only a single camera, similarly to \cite{gafni2020dynamic}.

Since \name{} uses the same set of inputs and outputs as those used in NeRF, it is applicable to many of NeRF's extensions. This allows for accelerating existing approaches for: reconstruction of dynamic scenes \cite{park2020deformable, dnerf, li2020neural}, single-image reconstruction \cite{tancik2020learned}, quality improvement \cite{rebain2020derf, zhang2020nerf++} and others \cite{neff2021donerf}. With minor modifications, \name{} can be applied to even more methods allowing for control over illumination \cite{neuralReflectanceFields} and incorporation of uncertainty \cite{nerfInTheWild}.

%% file: conclusion.tex
%!TEX root = egpaper_for_review.tex

\section{Conclusion}
In this paper we presented \name{}, a novel extension to NeRF that enables the rendering of photorealistic images at 200Hz and more on consumer hardware. We achieve significant speedups over NeRF and competing methods by factorizing its function approximator, enabling a caching step that makes rendering memory-bound instead of compute-bound. This allows for the application of NeRF to real-time scenarios.

%% file: supplementary/supplementary.tex
%%%%%%%%% BODY TEXT

\section{Overview}
We use the supplementary material to provide more detailed results and algorithms. We urge the reader to see our video, in which we show  results for all $16$ scenes we tested on along with providing an intuition for our method. On a practical note, we also provide guidance on getting smooth results for anyone adapting our method to their own work.

Please note that parameters chosen throughout this work for the rendering algorithm favour the highest possible quality. It is possible to significantly increase rendering speeds by sacrificing a small amount of visual quality for real-world applications.

\subsection{`Motion' Smoothness}
We notice that our method can exhibit more flickering artefacts on the NeRF 360 Synthetic dataset than the baseline \textit{when trained using the same settings as NeRF}, which we find only noticeable in motion. These direction-dependent artefacts are due to overfitting of the view-dependent MLP and can be successfully mitigated in one of two ways (which can also be combined). Sticking with identical parameters as NeRF, we can simply choose a resolution between $16^3$ and $64^3$ for the direction-dependent cache, which uses interpolation at lookup by default. Alternatively, we can adjust the Fourier feature encoding of the view directions at training time. In \cite{NeRF}, the Fourier encoding maps from $\mathbb{R}$ to $\mathbb{R}^{2L}$, where $L$ is $10$ for position, and $4$ for direction. For FastNeRF, setting $L=1$ actually works best for the NeRF 360 Synthetic dataset. Please note that we do not use these approaches when computing metrics in the text to ensure the same settings are used for all datasets, but that using them would slightly improve the results on synthetic data.

\textbf{Effect of modified Fourier feature encoding on metrics:} Using an $8$ layer $256$ hidden unit MLP for position, and a $4$ layer $128$ MLP for direction, we can obtain an average PSNR of $29.748dB$ for $L=1$, $29.646dB$ for $L=2$, $29.449dB$ for $L=3$ on synthetic data, with smoothness of results being reflected in these numbers.

\textbf{Effect of small direction cache on metrics:} Using a smaller direction cache has a negligible impact on metrics, occasionally actually improving them. This shows that the direction-dependent effects required by FastNeRF are low in frequency. For the scenes where we found we needed to use a smaller cache for moving images, the PSNR differences caused by using a smaller directional cache are: \textit{Materials} ($32^3$) (28.885dB \textrightarrow 28.874dB); \textit{Drums} ($16^3$) (23.745dB \textrightarrow 23.836dB); \textit{Lego} ($32^3$) (32.275dB \textrightarrow 32.155dB); \textit{Mic} ($32^3$) (31.765dB \textrightarrow 31.667dB); \textit{Hotdog} ($32^3$) (34.722dB \textrightarrow 34.644dB); \textit{Ficus} ($32^3$) (27.792dB \textrightarrow 28.193dB).

\subsection{Training \& Detailed Results}
For training, we use the same frequency encoding, noise perturbation and learning rate decay (starting at $5e-4$) as \cite{NeRF}. For the NeRF 360 Synthetic dataset, we use $64$ and $128$ samples for the coarse and fine networks, respectively. This becomes $64$ and $64$ for the LLFF scenes. Note that the coarse networks are discarded and not used in the caching stage - the cached density (or rather, the mesh derived from it) serves as the importance distribution during rendering. We sample $1000$ random rays per gradient descent step, and use Adam \cite{kingma2017adam} as our optimizer of choice, with $\beta_1=0.9$, $\beta_2=0.999$.

\begin{figure*}
    \centering
     \includegraphics[width=\textwidth]{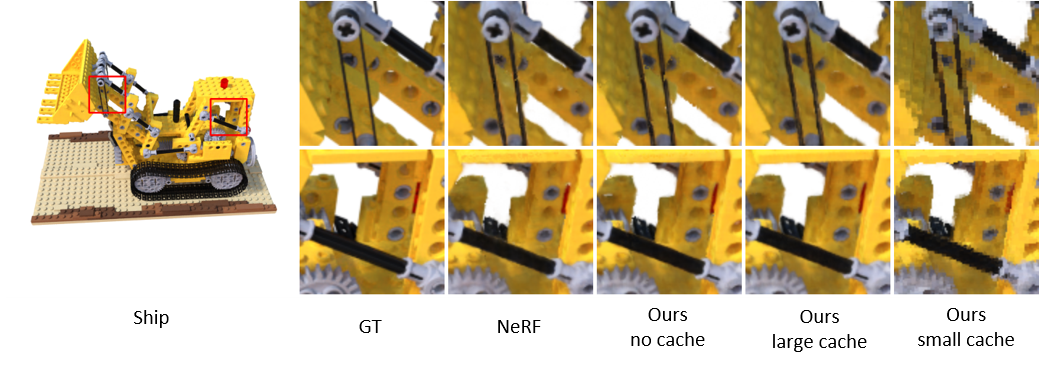}
     \caption{Qualitative comparison of our method vs NeRF on the \textit{Lego} scene from the dataset of \cite{NeRF} at $800^2$ pixels using 8 components. \textit{Small cache} refers to our method cached at $256^3$, and \textit{large cache} at $1024^3$.}
     \label{fig:qualitative_Lego}
\end{figure*}

In the paper, we compute test metrics on a random subset of 20 images per scene from the test sets of the NeRF 360 Synthetic dataset, and all available test images for the LLFF data. We use the evaluation set only to select the best iteration, which is 300K for all scenes, indicating identical convergence trends as NeRF \cite{NeRF}. For completeness, we show results for all scenes using the \textit{full} number of test images in Table~\ref{tab:SupplMainQuantitative}. We note that the results and trends are in agreement with the sub-sampled test set. Please see Figure~\ref{fig:qualitative_Lego} and beyond for a qualitative comparison in addition to the video. We include further ablations on the number of components in Table~\ref{tab:mainFactorsHorns} and Table~\ref{tab:mainFactorsFortress}.

\begin{table*}[]
\centering
\caption{As in the main text, we compare NeRF to our method when not cached to a grid, and when cached at a high resolution in terms of PSNR, SSIM and LPIPS, and also provide the average speed in ms of our method when cached. The first $8$ scenes comprise the dataset of \cite{NeRF} at $800^2$ pixels, the last $8$ scenes the LLFF dataset \cite{localLightfieldFusionPaper} at $504\times378$ pixels. We use $8$ components and a $1024^3$ cache for the NeRF Synthetic 360 scenes (except for the ship scene where we use $768^3$), and 6 components at $768^3$ for LLFF.}
\begin{tabular}{@{}lllllllllll@{}}
\toprule
\textbf{Scene} & \multicolumn{3}{c}{\textbf{NeRF}}              & \multicolumn{3}{c}{\textbf{Ours - No Cache}}   & \multicolumn{3}{c}{\textbf{Ours - Cache}} & \textbf{Speed} \\ \midrule
               & \textit{PSNR}\textuparrow & \textit{SSIM}\textuparrow & \textit{LPIPS}\textdownarrow & \textit{PSNR}\textuparrow & \textit{SSIM}\textuparrow & \textit{LPIPS}\textdownarrow & \textit{PSNR}\textuparrow   & \textit{SSIM}\textuparrow  & \textit{LPIPS}\textdownarrow & \\
\textbf{Drums} & $24.58$dB  & $0.92$  & $0.08$  & $23.868$dB  & $0.91$  & $0.084$  & $23.745$dB  & $0.913$  & $0.083$  & $4.5$ms \\ \midrule
\textbf{Ship} & $27.21$dB  & $0.83$  & $0.17$  & $27.241$dB  & $0.839$  & $0.155$  & $27.685$dB  & $0.805$  & $0.192$  & $4.9$ms \\ \midrule
\textbf{Mic} & $30.85$dB  & $0.97$  & $0.03$  & $30.274$dB  & $0.973$  & $0.023$  & $31.765$dB  & $0.977$  & $0.022$  & $2.6$ms \\ \midrule
\textbf{Ficus} & $28.86$dB  & $0.96$  & $0.03$  & $27.037$dB  & $0.948$  & $0.034$  & $27.792$dB  & $0.954$  & $0.031$  & $3.7$ms \\ \midrule
\textbf{Chair} & $31.16$dB  & $0.96$  & $0.04$  & $30.967$dB  & $0.96$  & $0.032$  & $32.322$dB  & $0.966$  & $0.032$  & $2.6$ms \\ \midrule
\textbf{Lego} & $30.41$dB  & $0.95$  & $0.03$  & $31.076$dB  & $0.955$  & $0.023$  & $32.275$dB  & $0.964$  & $0.022$  & $5.5$ms \\ \midrule
\textbf{Hotdog} & $34.7$dB  & $0.97$  & $0.03$  & $34.21$dB  & $0.97$  & $0.029$  & $34.722$dB  & $0.973$  & $0.031$  & $4.9$ms \\ \midrule
\textbf{Materials} & $28.93$dB  & $0.94$  & $0.03$  & $28.567$dB  & $0.943$  & $0.034$  & $28.885$dB  & $0.947$  & $0.034$  & $3.9$ms \\ \midrule
\textbf{Fern*} & $26.84$dB  & $0.86$  & $0.09$  & $26.325$dB  & $0.853$  & $0.105$  & $25.006$dB  & $0.822$  & $0.131$  & $1.7$ms \\ \midrule
\textbf{Flower*} & $28.32$dB  & $0.89$  & $0.06$  & $28.916$dB  & $0.912$  & $0.043$  & $27.012$dB  & $0.878$  & $0.059$  & $1.3$ms \\ \midrule
\textbf{Fortress*} & $32.7$dB  & $0.93$  & $0.03$  & $32.946$dB  & $0.937$  & $0.021$  & $31.256$dB  & $0.913$  & $0.043$  & $0.8$ms \\ \midrule
\textbf{Horns*} & $28.78$dB  & $0.9$  & $0.07$  & $29.748$dB  & $0.925$  & $0.044$  & $26.847$dB  & $0.889$  & $0.07$  & $1.2$ms \\ \midrule
\textbf{Leaves*} & $22.43$dB  & $0.82$  & $0.1$  & $22.53$dB  & $0.833$  & $0.095$  & $21.3$dB  & $0.787$  & $0.122$  & $1.5$ms \\ \midrule
\textbf{Orchids*} & $21.36$dB  & $0.75$  & $0.12$  & $21.204$dB  & $0.747$  & $0.126$  & $20.356$dB  & $0.712$  & $0.137$  & $1.6$ms \\ \midrule
\textbf{Room*} & $33.59$dB  & $0.96$  & $0.04$  & $33.702$dB  & $0.961$  & $0.034$  & $30.301$dB  & $0.942$  & $0.057$  & $1.6$ms \\ \midrule
\textbf{TRex*} & $27.74$dB  & $0.92$  & $0.05$  & $28.292$dB  & $0.933$  & $0.04$  & $26.204$dB  & $0.903$  & $0.063$  & $1.4$ms \\ \midrule
\end{tabular}
\label{tab:SupplMainQuantitative}
\end{table*}

\begin{table*}[]
\centering
\caption{Influence of the number of components and grid resolution on PSNR and memory required for caching the \textit{Horns} (`Triceratops') scene. Note how more factors also increase grid sparsity in this case. 6 components are a reasonable amount for LLFF data.}
\begin{tabular}{@{}l|ll|ll|ll|ll|ll@{}}
\toprule
\textbf{Factors} & \multicolumn{2}{c|}{\textbf{No Cache}} & \multicolumn{2}{c|}{\textbf{$256^3$}} & \multicolumn{2}{c|}{\textbf{$384^3$}} & \multicolumn{2}{c|}{\textbf{$512^3$}} & \multicolumn{2}{c}{\textbf{$768^3$}} \\
\textbf{}        & \textit{PSNR}\textuparrow    & \textit{Memory}   & \textit{PSNR}\textuparrow    & \textit{Memory}   & \textit{PSNR}\textuparrow    & \textit{Memory}   & \textit{PSNR}\textuparrow    & \textit{Memory}   & \textit{PSNR}\textuparrow             & \textit{Memory}           \\ \hline
\textbf{4} & $29.56$dB  & -  & $22.94$dB  & $0.39$Gb  & $24.64$dB  & $0.8$Gb  & $25.62$dB  & $1.56$Gb  & $26.52$dB  & $4.18$Gb \\
\textbf{6} & $29.75$dB  & -  & $23.08$dB  & $0.56$Gb  & $24.86$dB  & $1.14$Gb  & $25.89$dB  & $2.22$Gb  & $26.85$dB  & $6.11$Gb \\
\textbf{8} & $29.82$dB  & -  & $23.02$dB  & $0.72$Gb  & $24.79$dB  & $1.45$Gb  & $25.81$dB  & $2.81$Gb  & $26.76$dB  & $7.83$Gb \\
\end{tabular}
\label{tab:mainFactorsHorns}
\end{table*}

\begin{table*}[]
\centering
\caption{Influence of the number of components and grid resolution on PSNR and memory required for caching the \textit{Fortress} (`Minas Tirith') scene. Note how more factors also increase grid sparsity in this case. 6 components are a reasonable amount for LLFF data.}
\begin{tabular}{@{}l|ll|ll|ll|ll|ll@{}}
\toprule
\textbf{Factors} & \multicolumn{2}{c|}{\textbf{No Cache}} & \multicolumn{2}{c|}{\textbf{$256^3$}} & \multicolumn{2}{c|}{\textbf{$384^3$}} & \multicolumn{2}{c|}{\textbf{$512^3$}} & \multicolumn{2}{c}{\textbf{$768^3$}} \\
\textbf{}        & \textit{PSNR}\textuparrow    & \textit{Memory}   & \textit{PSNR}\textuparrow    & \textit{Memory}   & \textit{PSNR}\textuparrow    & \textit{Memory}   & \textit{PSNR}\textuparrow    & \textit{Memory}   & \textit{PSNR}\textuparrow             & \textit{Memory}           \\
\textbf{4} & $32.81$dB  & -  & $26.54$dB  & $0.42$Gb  & $28.61$dB  & $0.93$Gb  & $29.83$dB  & $1.88$Gb  & $30.99$dB  & $5.43$Gb \\
\textbf{6} & $32.95$dB  & -  & $26.63$dB  & $0.6$Gb  & $28.75$dB  & $1.31$Gb  & $30.01$dB  & $2.66$Gb  & $31.26$dB  & $7.83$Gb \\
\textbf{8} & $33.03$dB  & -  & $26.66$dB  & $0.79$Gb  & $28.81$dB  & $1.72$Gb  & $30.1$dB  & $3.47$Gb  & $31.36$dB  & $10.29$Gb \\
\end{tabular}
\label{tab:mainFactorsFortress}
\end{table*}

\subsection{Meshing \& Rendering}
One of the key advantages of working with sparse octrees to store the caches required by our method is that they can serve as a basis for accelerating the rendering process. Using raytracing, we can quickly terminate rays that miss the neurally rendered object(s) all together. For the rays that hit occupied voxels, and so require integrating the volume, we can use raytracing to skip empty space efficiently, saving a significant amount of queries. This matters because caching makes our method memory instead of compute bound. Computing the inner product of the components and weights as well as tracking transmittance for each ray is fast on modern GPUs. Grid look-ups, which require reading from the GPUs RAM on the other hand, are expensive.

While we could use a hierarchical digital differential analyzer such as \cite{ovdbHDDA} to determine ray grid intersections, we opt to trace rays against a collision mesh derived from the volume instead, for three reasons. First, this allows us to take advantage of hardware acceleration for the BVH and ray-triangle intersections on modern hardware. Second, we can down/resample or deform the meshes easily using existing tools. Finally, meshes could be used to `fake' shadows, bouncelight and other effects which can be composited over the neurally rendered content.

\begin{figure*}
    \centering
     \includegraphics[width=\textwidth]{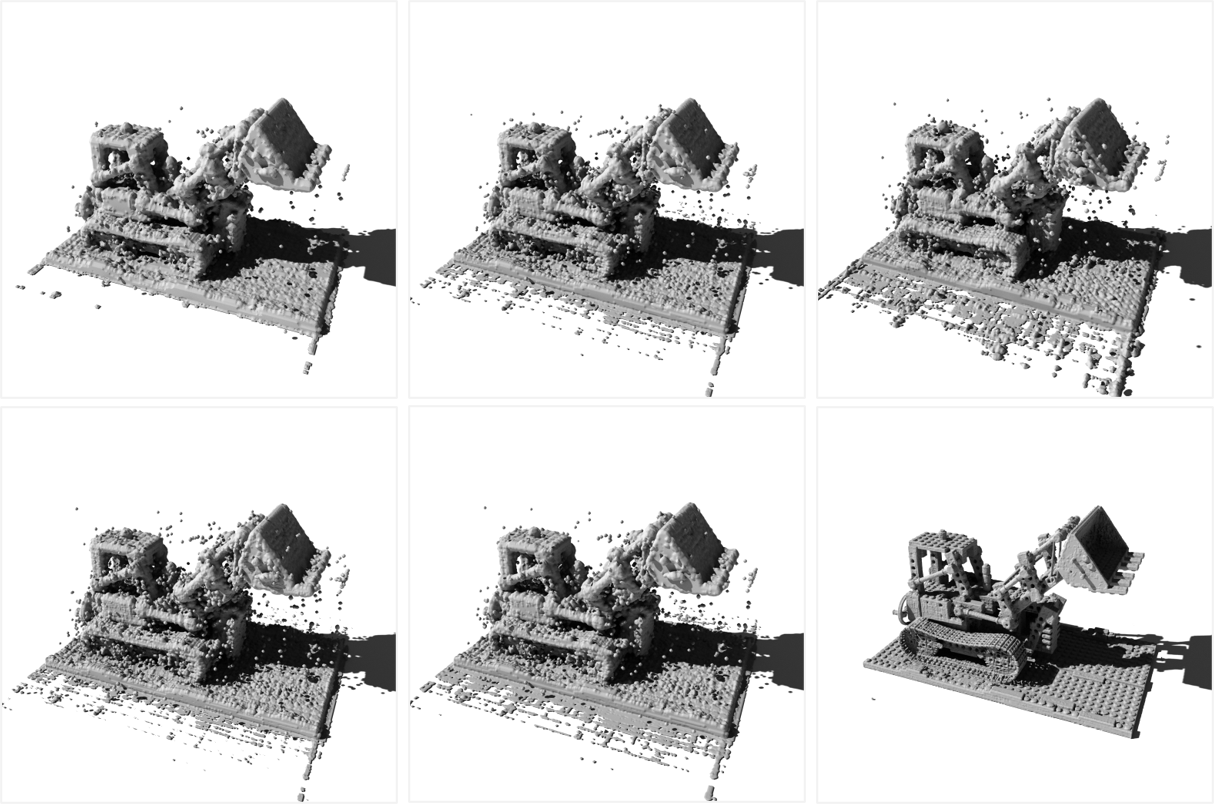}
     \caption{Collision Meshes for the \textit{Lego} scene used for our raytraced rendering algorithm. From left to right, top to bottom, these are derived from the following grid resolutions $256^3$, $384^3$, $512^3$, $768^3$, $1024^3$. Note that we threshold the density at $0$, but being less conservative can give meshes good enough to compute shadows or bounce light. This is seen in the bottom-right image, where the identical volume is thresholded at $10.0$ instead (best viewed zoomed in).}
     \label{fig:meshing}
\end{figure*}

We derive the meshes by converting the density cache to a sign distance function (thresholding at $0.0$), optionally downsampling it before meshing with marching cubes, and optionally simplifying the resulting geometry using standard techniques. We show the resulting geometry for the \textit{Lego} scene in Figure~\ref{fig:meshing}. Note that more aggressive thresholding of the volume or remeshing is easily possible and can lead to significant performance benefits in practise as more rays can be culled early and mesh complexity reduced.

For rendering, we follow the same method as NeRF, using the Beer-Lambert Law \cite{FongVolumeRendering} to model radiance extinction. Once the contribution of new samples for a ray reaches $0.001$, we terminate a ray's rendering kernel. Note that this can be relaxed for greater performance. Another way to accelerate the rendering process is to visit fewer voxels based on radiance or transmittance. For all paper experiments however, we never skip occupied space, or optimise any parameters per scene.

\subsection{Cache size calculation}
In this section we show the formulas we used to estimate cache sizes $M_{NeRF}, M_{FastNeRF}$ for NeRF and FastNeRF respectively.
\begin{equation}
    M_{NeRF} = \alpha(s_{\sigma} + s_{rgb})k^3l^2,
\end{equation}
where $s_{\sigma}, s_{rgb}$ are the sizes of the stored transparency and RGB values in bits and $\alpha \in \langle 0, 1)$ is the inverse volume sparsity, where $\alpha=1$ would indicate a dense volume.
\begin{equation}
     M_{FastNeRF} = \alpha \left( (D \cdot s_{abc} + s_{\sigma})k^3 \right) + D \cdot s_{\beta} l^2,
\end{equation}
where $s_{uvw}, s_{\beta}$ are the sizes of the stored $(u_i,v_i,w_i)$ values and the weights $\beta_i$.

For $k=l=1024$, $(r,g,b)$ stored as individual bytes and all other values stored as half-precision floats  the cache size would be
\begin{equation}
    M_{NeRF} = \alpha \cdot 5,629,499,534,213,120 \approx \alpha \cdot 5.6PB,
\end{equation}
for NeRF and
\begin{equation}
    M_{factor}=\alpha \cdot 105,226,698,752 + 33,554,532 \approx \alpha \cdot 105GB,
\end{equation}
for FastNeRF.

Even for highly sparse volumes, where $\alpha=0.01$, the cache would take nearly 60TB, which exceeds the memory capacity of consumer-grade hardware. The high memory requirement makes this approach impractical when used with the standard NeRF architecture.

\subsection{Cache size calculation}
In this section we show the formulas we used to estimate cache sizes $M_{NeRF}, M_{FastNeRF}$ for NeRF and FastNeRF respectively.
\begin{equation}
    M_{NeRF} = \alpha(s_{\sigma} + s_{rgb})k^3l^2,
\end{equation}
where $s_{\sigma}, s_{rgb}$ are the sizes of the stored transparency and RGB values in bits and $\alpha \in \langle 0, 1 \rangle$ is the inverse volume sparsity, where $\alpha=1$ would indicate a dense volume. Just as in the main paper $k$, $l$ correspond to the number of bins per dimension in the position and direction dependent caches respectively.
\begin{equation}
     M_{FastNeRF} = \alpha \left( (D \cdot s_{uvw} + s_{\sigma})k^3 \right) + D \cdot s_{\beta} l^2,
\end{equation}
where $s_{uvw}, s_{\beta}$ are the sizes of the stored $(u_i,v_i,w_i)$ values and the weights $\beta_i$.

For $k=l=1024, D=8$, $(r,g,b)$ stored as individual bytes and all other values stored as half-precision floats the cache size would be
\begin{equation}
    M_{NeRF} = \alpha \cdot 5,629,499,534,213,120 \approx \alpha \cdot 5.6PB,
\end{equation}
for NeRF and
\begin{equation}
    M_{factor}=\alpha \cdot 53,687,091,200 + 16,777,216 \approx \alpha \cdot 54GB,
\end{equation}
for FastNeRF.

\subsection{Details on Applications}
In our proof-of-concept telepresence scenario we use a deformation field network \cite{park2020deformable} $\mathcal{F}_{deform}$  that modifies the input sample positions. This network is similar to the position-dependent network used in FastNeRF, but smaller - a 6-layer MLP with 64 units in each layer. The input is a sample position and an expression vector. The sample position is processed with positional encoding identically to how the FastNeRF input position is processed. The output is an offset that is applied to the input sample position. When training with $\mathcal{F}_{deform}$ we add an L2 regularizer that constrains $\mathcal{F}_{deform}$ to be an identity transform if a neutral expression is passed as input. We find that this solution stabilizes training and improves results.

In addition to $\mathcal{F}_{deform}$, we also use a non-trainable deformation field $\mathcal{F}_{bone}$ that moves the samples in the head region according to the movement of the 3D model bones that represent the shoulders and the neck. This additional deformation accounts for large movements of the head and the shoulders, which reduces the load on $\mathcal{F}_{deform}$ and improves quality.
The two deformation fields are composed with the position-dependent network $\mathcal{F}_{pos}$ as follows:
\begin{equation}
    \{ \sigma, (\bm{u}, \bm{v}, \bm{w}) \} = \mathcal{F}_{pos}
    \left(\mathcal{F}_{deform}(\mathcal{F}_{bone}(\bm{p}, \bm{e}), \bm{e})\right),
\end{equation}
where $\bm{e}$ is the expression vector. While we could use $\mathcal{F}_{bone}$ at test-time as well, we remove it to improve performance.

The training procedure in this scenario is identical to that used in FastNeRF but with 64 samples in the coarse stage and an additional 64 in the fine stage. The position-dependent network has 384 units in each layer, while the view-dependent network uses only 32 units and 2 layers. The view-dependent network is very small as we believe the scene's illumination is simple and fairly uniform.

The main limitation of this approach in terms of speed is the need to call $\mathcal{F}_{deform}$ for all the input samples. In order to mitigate this, at test-time we implement a grid-based sample pruner. For each position in a grid around the face region we compute the density $\sigma$
\begin{equation}
 \sigma = \mathcal{F}_{pos}\left( \mathcal{F}_{deform}(\bm{p}, \bm{e}) \right)
\end{equation}
for 50 randomly chosen samples of $\bm{e}$. For each grid position we record the minimum and maximum encountered $\sigma$ in a sparse volume.  At test-time we use this volume to prune the inputs to $\mathcal{F}_{deform}$ where the maximum $\sigma$ is 0 and where the rays would get saturated, which we estimate using the minimum density values from the volume. This approach reduces the runtime of $\mathcal{F}_{deform}$ by more than a factor of 2.

To further reduce the impact of $\mathcal{F}_{deform}$ on the runtime we only use 43 samples in the coarse stage and remove the fine stage altogether at test time. To offset this low sample count we add additional samples that are linearly interpolated between the samples generated by the deformation network
\begin{equation}
    \bm{p}_{deform}=\mathcal{F}_{deform}(\bm{p}, \bm{e})).
\end{equation}
The additional samples can be evaluated very cheaply as they can be looked-up in the FastNeRF cache. Note that since the deformation network changes the positions of individual samples along the rays, the rays are no longer straight. This means that in this scenario we cannot use the hardware-accelerated ray tracing procedure described in the main paper, though the pruning method described above serves a similar role.

Even though the steps above significantly reduce the runtime of $\mathcal{F}_{deform}$, we are still limited to rendering $300\times300$ pixel images with a reduced sample count if we want to maintain 30FPS with the deformation network. We observed that if this framerate constraint was to be disregarded, the approach described above is able to generate images at significantly better quality. Thus, we believe that a faster deformation approach remains an important goal for future work.

\begin{figure*}
    \centering
     \includegraphics[width=\textwidth]{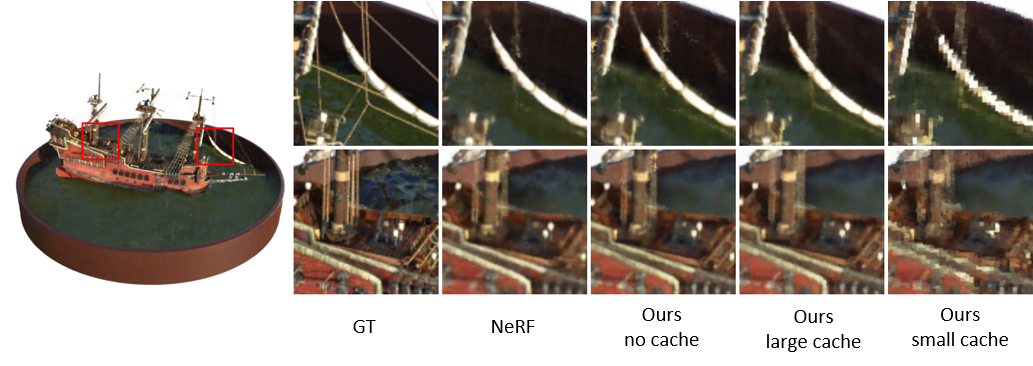}
     \caption{Qualitative comparison of our method vs NeRF on the \textit{Ship} scene from the dataset of \cite{NeRF} at $800^2$ pixels using 8 components. \textit{Small cache} refers to our method cached at $256^3$, and \textit{large cache} at $1024^3$.}
     \label{fig:qualitative_Ship}
\end{figure*}
\begin{figure*}
    \centering
     \includegraphics[width=\textwidth]{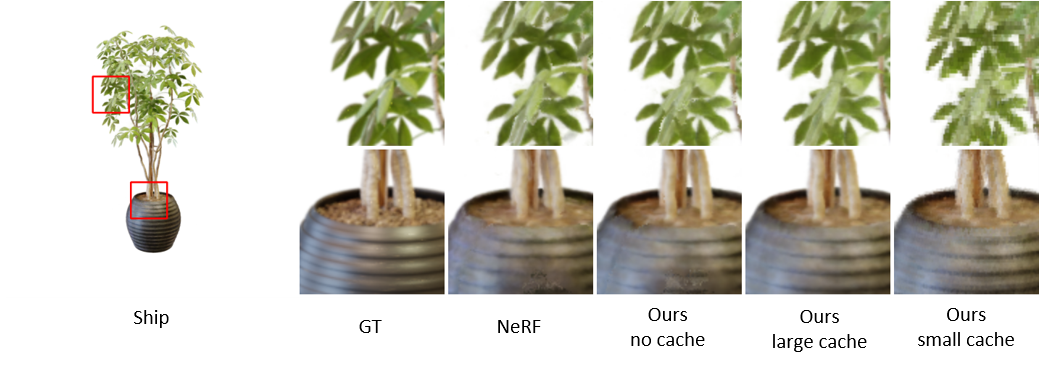}
     \caption{Qualitative comparison of our method vs NeRF on the \textit{Ficus} scene from the dataset of \cite{NeRF} at $800^2$ pixels using 8 components. \textit{Small cache} refers to our method cached at $256^3$, and \textit{large cache} at $1024^3$.}
     \label{fig:qualitative_Ficus}
\end{figure*}
\begin{figure*}
    \centering
     \includegraphics[width=\textwidth]{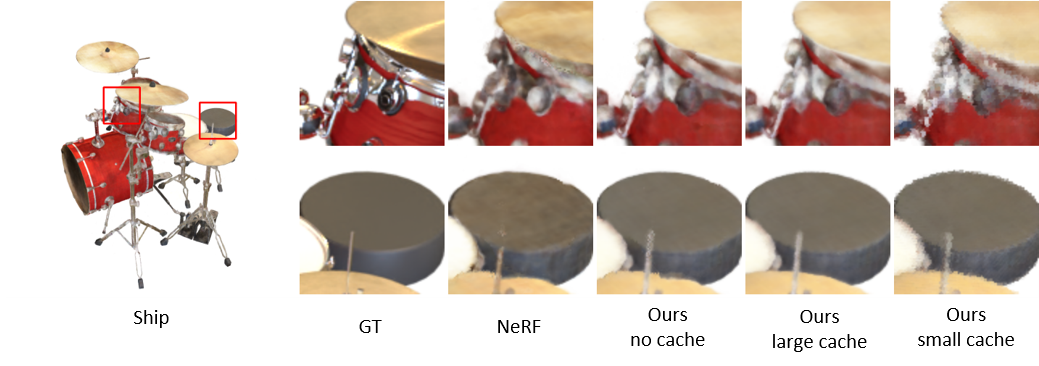}
     \caption{Qualitative comparison of our method vs NeRF on the \textit{Drums} scene from the dataset of \cite{NeRF} at $800^2$ pixels using 8 components. \textit{Small cache} refers to our method cached at $256^3$, and \textit{large cache} at $1024^3$.}
     \label{fig:qualitative_Drums}
\end{figure*}
\begin{figure*}
    \centering
     \includegraphics[width=\textwidth]{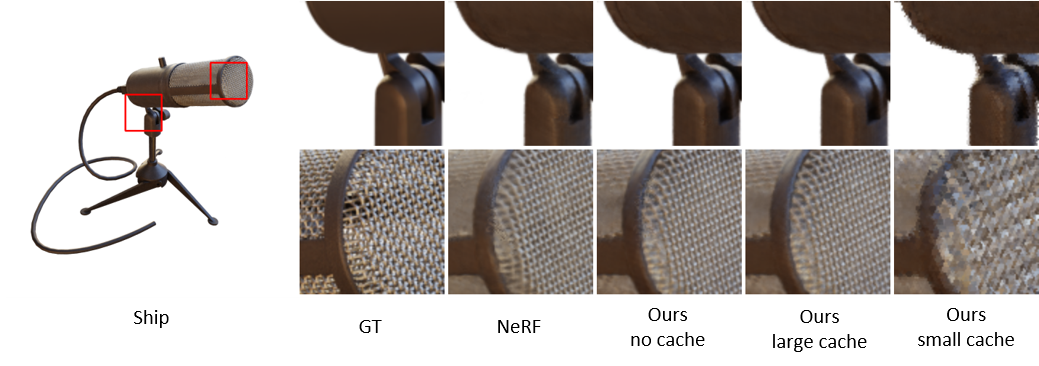}
     \caption{Qualitative comparison of our method vs NeRF on the \textit{Mic} scene from the dataset of \cite{NeRF} at $800^2$ pixels using 8 components. \textit{Small cache} refers to our method cached at $256^3$, and \textit{large cache} at $1024^3$.}
     \label{fig:qualitative_Mic}
\end{figure*}
\begin{figure*}
    \centering
     \includegraphics[width=\textwidth]{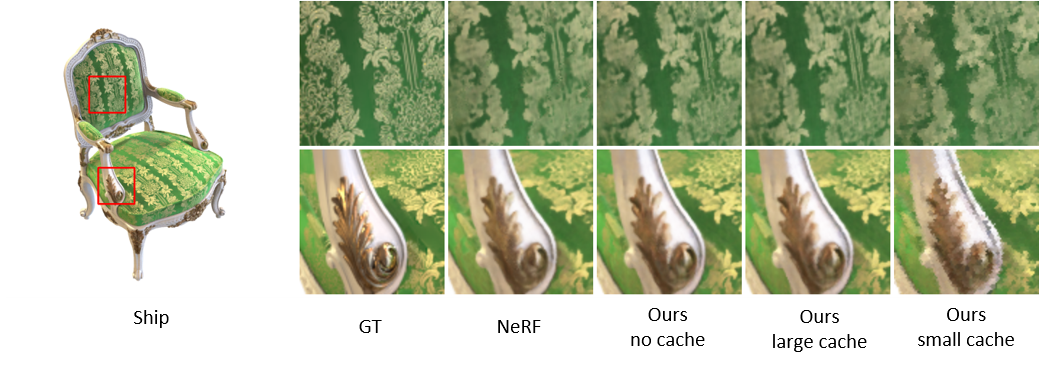}
     \caption{Qualitative comparison of our method vs NeRF on the \textit{Chair} scene from the dataset of \cite{NeRF} at $800^2$ pixels using 8 components. \textit{Small cache} refers to our method cached at $256^3$, and \textit{large cache} at $1024^3$.}
     \label{fig:qualitative_Chair}
\end{figure*}
\begin{figure*}
    \centering
     \includegraphics[width=\textwidth]{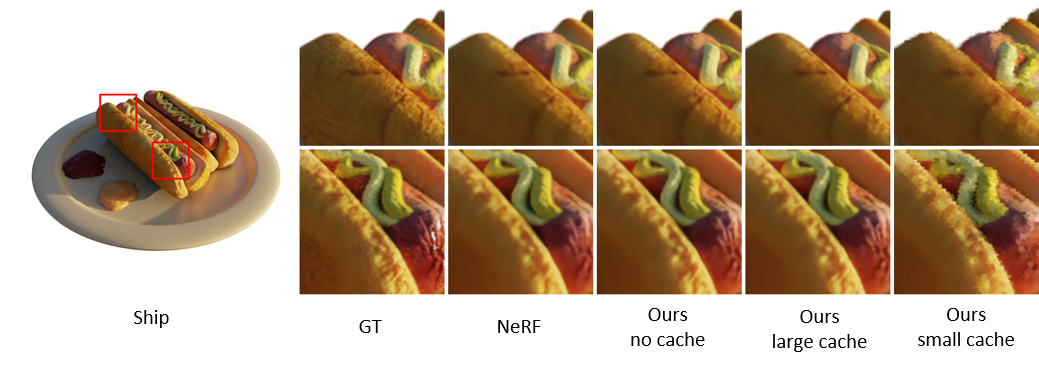}
     \caption{Qualitative comparison of our method vs NeRF on the \textit{Hotdog} scene from the dataset of \cite{NeRF} at $800^2$ pixels using 8 components. \textit{Small cache} refers to our method cached at $256^3$, and \textit{large cache} at $1024^3$.}
     \label{fig:qualitative_Hotdog}
\end{figure*}
\begin{figure*}
    \centering
     \includegraphics[width=\textwidth]{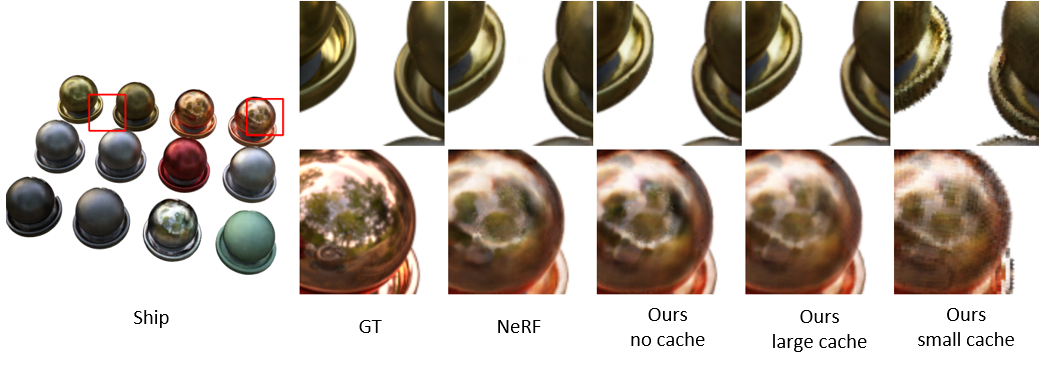}
     \caption{Qualitative comparison of our method vs NeRF on the \textit{Materials} scene from the dataset of \cite{NeRF} at $800^2$ pixels using 8 components. \textit{Small cache} refers to our method cached at $256^3$, and \textit{large cache} at $1024^3$.}
     \label{fig:qualitative_Materials}
\end{figure*}

\begin{figure*}
    \centering
     \includegraphics[width=\textwidth]{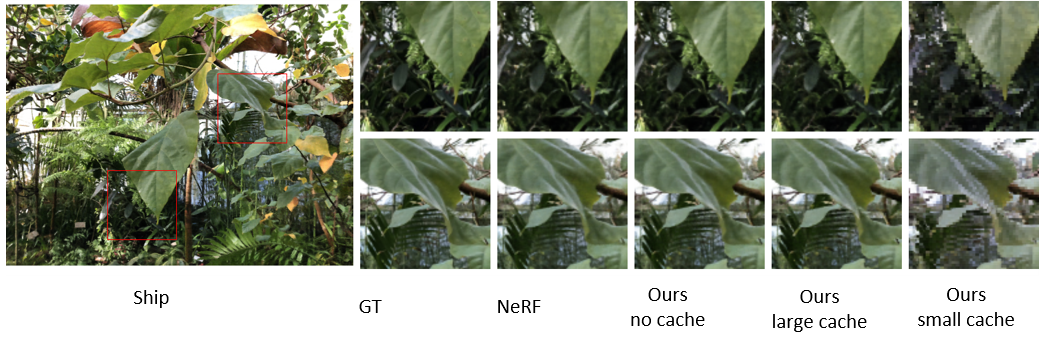}
     \caption{Qualitative comparison of our method vs NeRF on the \textit{Leaves} scene from the dataset of \cite{localLightfieldFusionPaper} at $504\times378$ pixels using 6 factors. \textit{Small cache} refers to our method cached at $256^3$, and \textit{large cache} at $768^3$.}
     \label{fig:qualitative_Leaves}
\end{figure*}

\begin{figure*}
    \centering
     \includegraphics[width=\textwidth]{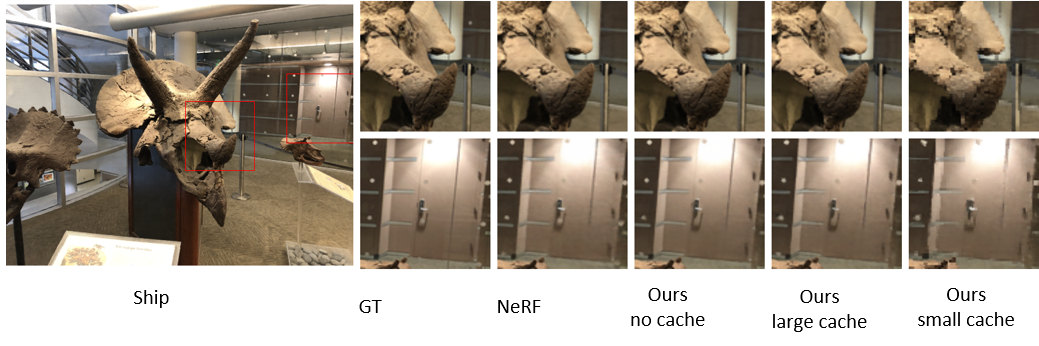}
     \caption{Qualitative comparison of our method vs NeRF on the \textit{Horns} scene from the dataset of \cite{localLightfieldFusionPaper} at $504\times378$ pixels using 6 factors. \textit{Small cache} refers to our method cached at $256^3$, and \textit{large cache} at $768^3$.}
     \label{fig:qualitative_Horns}
\end{figure*}

\begin{figure*}
    \centering
     \includegraphics[width=\textwidth]{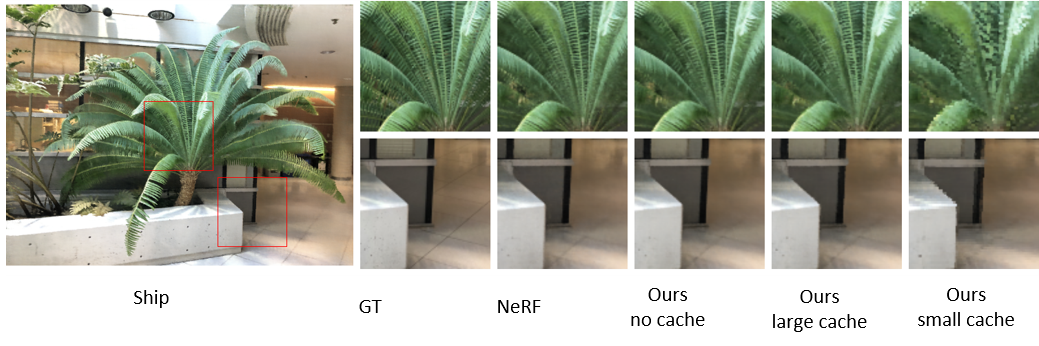}
     \caption{Qualitative comparison of our method vs NeRF on the \textit{Fern} scene from the dataset of \cite{localLightfieldFusionPaper} at $504\times378$ pixels using 6 factors. \textit{Small cache} refers to our method cached at $256^3$, and \textit{large cache} at $768^3$.}
     \label{fig:qualitative_Fern}
\end{figure*}

\begin{figure*}
    \centering
     \includegraphics[width=\textwidth]{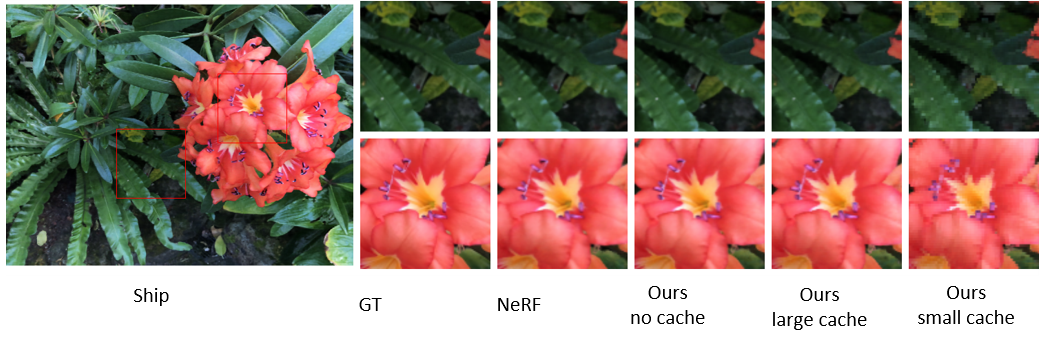}
     \caption{Qualitative comparison of our method vs NeRF on the \textit{Flower} scene from the dataset of \cite{localLightfieldFusionPaper} at $504\times378$ pixels using 6 factors. \textit{Small cache} refers to our method cached at $256^3$, and \textit{large cache} at $768^3$.}
     \label{fig:qualitative_Flower}
\end{figure*}

\begin{figure*}
    \centering
     \includegraphics[width=\textwidth]{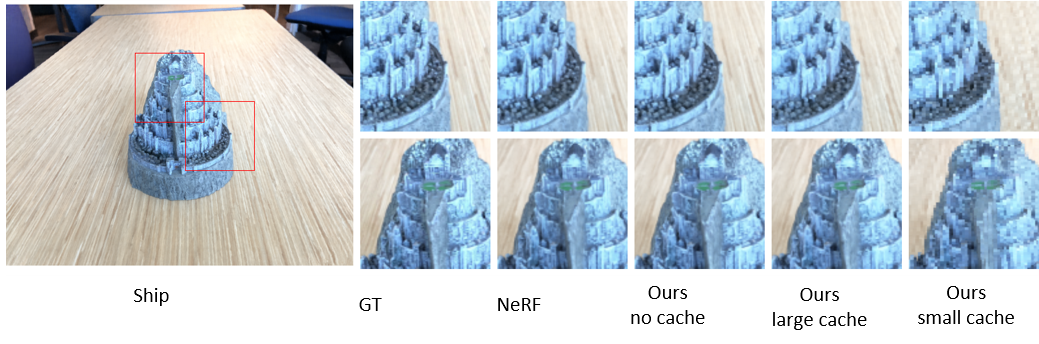}
     \caption{Qualitative comparison of our method vs NeRF on the \textit{Fortress} (`Minas Tirith') scene from the dataset of \cite{localLightfieldFusionPaper} at $504\times378$ pixels using 6 factors. \textit{Small cache} refers to our method cached at $256^3$, and \textit{large cache} at $768^3$.}
     \label{fig:qualitative_Fortress}
\end{figure*}

\begin{figure*}
    \centering
     \includegraphics[width=\textwidth]{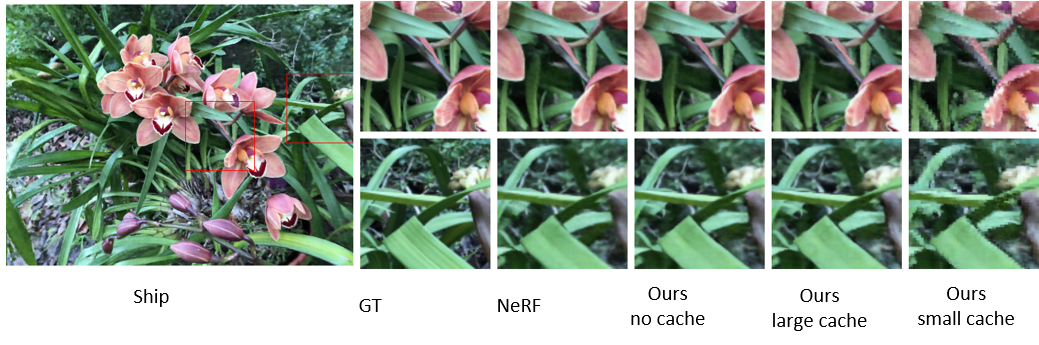}
     \caption{Qualitative comparison of our method vs NeRF on the \textit{Orchids} scene from the dataset of \cite{localLightfieldFusionPaper} at $504\times378$ pixels using 6 factors. \textit{Small cache} refers to our method cached at $256^3$, and \textit{large cache} at $768^3$.}
     \label{fig:qualitative_Orchids}
\end{figure*}

\begin{figure*}
    \centering
     \includegraphics[width=\textwidth]{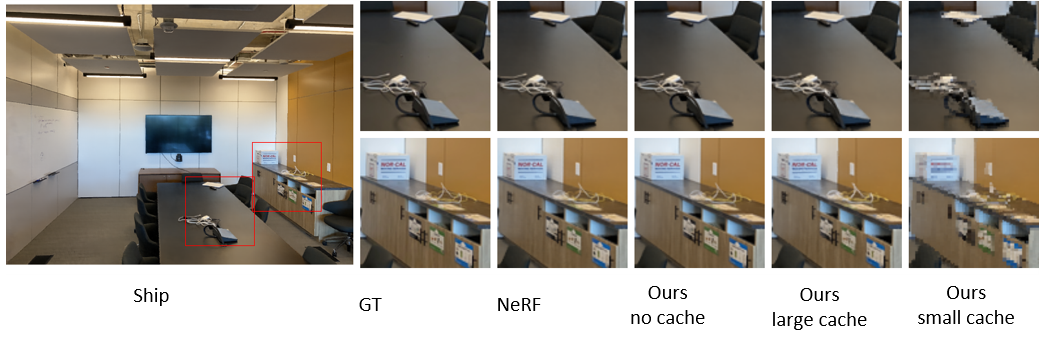}
     \caption{Qualitative comparison of our method vs NeRF on the \textit{Room} scene from the dataset of \cite{localLightfieldFusionPaper} at $504\times378$ pixels using 6 factors. \textit{Small cache} refers to our method cached at $256^3$, and \textit{large cache} at $768^3$.}
     \label{fig:qualitative_Room}
\end{figure*}

\begin{figure*}
    \centering
     \includegraphics[width=\textwidth]{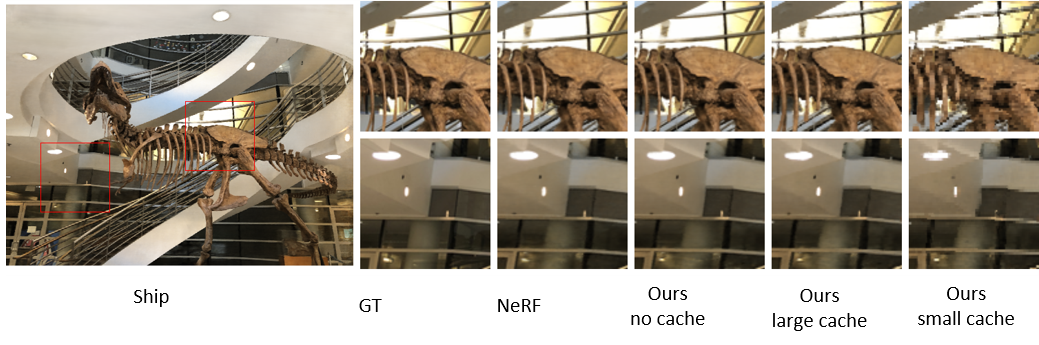}
     \caption{Qualitative comparison of our method vs NeRF on the \textit{TRex} scene from the dataset of \cite{localLightfieldFusionPaper} at $504\times378$ pixels using 6 factors. \textit{Small cache} refers to our method cached at $256^3$, and \textit{large cache} at $768^3$.}
     \label{fig:qualitative_TRex}
\end{figure*}

%% file: egpaper_for_review.bbl
\begin{thebibliography}{10}\itemsep=-1pt

\bibitem{xfields}
Mojtaba Bemana, Karol Myszkowski, Hans-Peter Seidel, and Tobias Ritschel.
\newblock X-fields: Implicit neural view-, light- and time-image interpolation.
\newblock {\em ACM Transactions on Graphics (Proc. SIGGRAPH Asia 2020)}, 39(6),
  2020.

\bibitem{neuralReflectanceFields}
Sai Bi, Zexiang Xu, Pratul Srinivasan, Ben Mildenhall, Kalyan Sunkavalli,
  Miloš Hašan, Yannick Hold-Geoffroy, David Kriegman, and Ravi Ramamoorthi.
\newblock Neural reflectance fields for appearance acquisition, 2020.

\bibitem{deepReflectanceVolumes}
Sai Bi, Zexiang Xu, Kalyan Sunkavalli, Miloš Hašan, Yannick Hold-Geoffroy,
  David Kriegman, and Ravi Ramamoorthi.
\newblock Deep reflectance volumes: Relightable reconstructions from multi-view
  photometric images, 2020.

\bibitem{originalSphericalHarmonics}
Brian Cabral, Nelson Max, and Rebecca Springmeyer.
\newblock Bidirectional reflection functions from surface bump maps.
\newblock 21(4):273–281, Aug. 1987.

\bibitem{implicitFunctionsShapeModelling}
Zhiqin Chen and Hao Zhang.
\newblock Learning implicit fields for generative shape modeling.
\newblock {\em CoRR}, abs/1812.02822, 2018.

\bibitem{jaxnerf2020github}
Boyang Deng, Jonathan~T. Barron, and Pratul~P. Srinivasan.
\newblock {JaxNeRF}: an efficient {JAX} implementation of {NeRF}, 2020.

\bibitem{FongVolumeRendering}
Julian Fong, Magnus Wrenninge, Christopher Kulla, and Ralf Habel.
\newblock Production volume rendering: Siggraph 2017 course.
\newblock In {\em ACM SIGGRAPH 2017 Courses}, SIGGRAPH '17, New York, NY, USA,
  2017. Association for Computing Machinery.

\bibitem{gafni2020dynamic}
Guy Gafni, Justus Thies, Michael Zollh{\"o}fer, and Matthias Nie{\ss}ner.
\newblock Dynamic neural radiance fields for monocular 4d facial avatar
  reconstruction.
\newblock {\em arXiv preprint arXiv:2012.03065}, 2020.

\bibitem{VCMPaper}
Iliyan Georgiev, Jaroslav K{\v{r}}iv{\'{a}}nek, Tom{\'{a}}{\v{s}}
  Davidovi{\v{c}}, and Philipp Slusallek.
\newblock Light transport simulation with vertex connection and merging.
\newblock {\em ACM Trans. Graph.}, 31(6):192:1--192:10, Nov. 2012.

\bibitem{pathTracingPaper}
James~T. Kajiya.
\newblock The rendering equation.
\newblock {\em SIGGRAPH Comput. Graph.}, 20(4):143–150, Aug. 1986.

\bibitem{styleganPaper}
Tero Karras, Samuli Laine, and Timo Aila.
\newblock A style-based generator architecture for generative adversarial
  networks, 2019.

\bibitem{kingma2017adam}
Diederik~P. Kingma and Jimmy Ba.
\newblock Adam: A method for stochastic optimization, 2017.

\bibitem{kolos2020transpr}
Maria Kolos, Artem Sevastopolsky, and Victor Lempitsky.
\newblock Transpr: Transparency ray-accumulating neural 3d scene point
  renderer, 2020.

\bibitem{lassner2020pulsar}
Christoph Lassner and Michael Zollhöfer.
\newblock Pulsar: Efficient sphere-based neural rendering, 2020.

\bibitem{flame}
Tianye Li, Timo Bolkart, Michael~J Black, Hao Li, and Javier Romero.
\newblock Learning a model of facial shape and expression from 4d scans.
\newblock {\em ACM Trans. Graph.}, 36(6):194--1, 2017.

\bibitem{li2020neural}
Zhengqi Li, Simon Niklaus, Noah Snavely, and Oliver Wang.
\newblock Neural scene flow fields for space-time view synthesis of dynamic
  scenes.
\newblock {\em arXiv preprint arXiv:2011.13084}, 2020.

\bibitem{lindell2020autoint}
David~B Lindell, Julien~NP Martel, and Gordon Wetzstein.
\newblock Autoint: Automatic integration for fast neural volume rendering.
\newblock {\em arXiv preprint arXiv:2012.01714}, 2020.

\bibitem{liu2021neural}
Lingjie Liu, Jiatao Gu, Kyaw~Zaw Lin, Tat-Seng Chua, and Christian Theobalt.
\newblock Neural sparse voxel fields, 2021.

\bibitem{codecAvatarsPaper}
Stephen Lombardi, Jason~M. Saragih, Tomas Simon, and Yaser Sheikh.
\newblock Deep appearance models for face rendering.
\newblock {\em CoRR}, abs/1808.00362, 2018.

\bibitem{NeuralVolumes}
Stephen Lombardi, Tomas Simon, Jason Saragih, Gabriel Schwartz, Andreas
  Lehrmann, and Yaser Sheikh.
\newblock Neural volumes: Learning dynamic renderable volumes from images.
\newblock {\em ACM Trans. Graph.}, 38(4):65:1--65:14, July 2019.

\bibitem{lombardi2021mixture}
Stephen Lombardi, Tomas Simon, Gabriel Schwartz, Michael Zollhoefer, Yaser
  Sheikh, and Jason Saragih.
\newblock Mixture of volumetric primitives for efficient neural rendering,
  2021.

\bibitem{marchingCubes}
William~E. Lorensen and Harvey~E. Cline.
\newblock Marching cubes: A high resolution 3d surface construction algorithm.
\newblock {\em SIGGRAPH Comput. Graph.}, 21(4):163–169, Aug. 1987.

\bibitem{loDBook}
David Luebke, Martin Reddy, Jonathan~D. Cohen, Amitabh Varshney, Benjamin
  Watson, and Robert Huebner.
\newblock {\em Level of Detail for 3D Graphics}.
\newblock Morgan Kaufmann Publishers Inc., San Francisco, CA, USA, 2002.

\bibitem{nerfInTheWild}
Ricardo Martin-Brualla, Noha Radwan, Mehdi S.~M. Sajjadi, Jonathan~T. Barron,
  Alexey Dosovitskiy, and Daniel Duckworth.
\newblock Nerf in the wild: Neural radiance fields for unconstrained photo
  collections, 2021.

\bibitem{occupancyNetworks}
Lars~M. Mescheder, Michael Oechsle, Michael Niemeyer, Sebastian Nowozin, and
  Andreas Geiger.
\newblock Occupancy networks: Learning 3d reconstruction in function space.
\newblock {\em CoRR}, abs/1812.03828, 2018.

\bibitem{localLightfieldFusionPaper}
Ben Mildenhall, Pratul~P. Srinivasan, Rodrigo Ortiz-Cayon, Nima~Khademi
  Kalantari, Ravi Ramamoorthi, Ren Ng, and Abhishek Kar.
\newblock Local light field fusion: Practical view synthesis with prescriptive
  sampling guidelines.
\newblock {\em ACM Trans. Graph.}, 38(4), July 2019.

\bibitem{NeRF}
Ben Mildenhall, Pratul~P Srinivasan, Matthew Tancik, Jonathan~T Barron, Ravi
  Ramamoorthi, and Ren Ng.
\newblock Nerf: Representing scenes as neural radiance fields for view
  synthesis.
\newblock {\em arXiv preprint arXiv:2003.08934}, 2020.

\bibitem{ovdbHDDA}
Ken Museth.
\newblock Hierarchical digital differential analyzer for efficient ray-marching
  in openvdb.
\newblock In {\em ACM SIGGRAPH 2014 Talks}, SIGGRAPH '14, New York, NY, USA,
  2014. Association for Computing Machinery.

\bibitem{deepshading}
Oliver Nalbach, Elena Arabadzhiyska, Dushyant Mehta, H-P Seidel, and Tobias
  Ritschel.
\newblock Deep shading: convolutional neural networks for screen space shading.
\newblock In {\em Computer graphics forum}, volume~36, pages 65--78. Wiley
  Online Library, 2017.

\bibitem{neff2021donerf}
Thomas Neff, Pascal Stadlbauer, Mathias Parger, Andreas Kurz, Chakravarty
  R.~Alla Chaitanya, Anton Kaplanyan, and Markus Steinberger.
\newblock Donerf: Towards real-time rendering of neural radiance fields using
  depth oracle networks, 2021.

\bibitem{deepSDF}
Jeong~Joon Park, Peter Florence, Julian Straub, Richard~A. Newcombe, and Steven
  Lovegrove.
\newblock Deepsdf: Learning continuous signed distance functions for shape
  representation.
\newblock {\em CoRR}, abs/1901.05103, 2019.

\bibitem{park2020deformable}
Keunhong Park, Utkarsh Sinha, Jonathan~T Barron, Sofien Bouaziz, Dan~B Goldman,
  Steven~M Seitz, and Ricardo-Martin Brualla.
\newblock Deformable neural radiance fields.
\newblock {\em arXiv preprint arXiv:2011.12948}, 2020.

\bibitem{nerfies}
Keunhong Park, Utkarsh Sinha, Jonathan~T. Barron, Sofien Bouaziz, Dan~B
  Goldman, Steven~M. Seitz, and Ricardo Martin-Brualla.
\newblock Deformable neural radiance fields.
\newblock {\em arXiv preprint arXiv:2011.12948}, 2020.

\bibitem{parker2010optix}
Steven~G Parker, James Bigler, Andreas Dietrich, Heiko Friedrich, Jared
  Hoberock, David Luebke, David McAllister, Morgan McGuire, Keith Morley,
  Austin Robison, et~al.
\newblock Optix: a general purpose ray tracing engine.
\newblock {\em Acm transactions on graphics (tog)}, 29(4):1--13, 2010.

\bibitem{pharr2016physically}
M. Pharr, W. Jakob, and G. Humphreys.
\newblock {\em Physically Based Rendering: From Theory to Implementation}.
\newblock Elsevier Science, 2016.

\bibitem{dnerf}
Albert Pumarola, Enric Corona, Gerard Pons-Moll, and Francesc Moreno-Noguer.
\newblock D-nerf: Neural radiance fields for dynamic scenes.
\newblock {\em arXiv preprint arXiv:2011.13961}, 2020.

\bibitem{rebain2020derf}
Daniel Rebain, Wei Jiang, Soroosh Yazdani, Ke Li, Kwang~Moo Yi, and Andrea
  Tagliasacchi.
\newblock Derf: Decomposed radiance fields.
\newblock {\em arXiv preprint arXiv:2011.12490}, 2020.

\bibitem{stableViewSynthesis}
Gernot Riegler and Vladlen Koltun.
\newblock Stable view synthesis, 2020.

\bibitem{srinivasan2020nerv}
Pratul~P. Srinivasan, Boyang Deng, Xiuming Zhang, Matthew Tancik, Ben
  Mildenhall, and Jonathan~T. Barron.
\newblock Nerv: Neural reflectance and visibility fields for relighting and
  view synthesis, 2020.

\bibitem{takikawa2021neural}
Towaki Takikawa, Joey Litalien, Kangxue Yin, Karsten Kreis, Charles Loop, Derek
  Nowrouzezahrai, Alec Jacobson, Morgan McGuire, and Sanja Fidler.
\newblock Neural geometric level of detail: Real-time rendering with implicit
  3d shapes, 2021.

\bibitem{tancik2020learned}
Matthew Tancik, Ben Mildenhall, Terrance Wang, Divi Schmidt, Pratul~P.
  Srinivasan, Jonathan~T. Barron, and Ren Ng.
\newblock Learned initializations for optimizing coordinate-based neural
  representations, 2020.

\bibitem{tancik2020fourier}
Matthew Tancik, Pratul~P. Srinivasan, Ben Mildenhall, Sara Fridovich-Keil,
  Nithin Raghavan, Utkarsh Singhal, Ravi Ramamoorthi, Jonathan~T. Barron, and
  Ren Ng.
\newblock Fourier features let networks learn high frequency functions in low
  dimensional domains, 2020.

\bibitem{NIPS2017_3f5ee243}
Ashish Vaswani, Noam Shazeer, Niki Parmar, Jakob Uszkoreit, Llion Jones,
  Aidan~N Gomez, \L~ukasz Kaiser, and Illia Polosukhin.
\newblock Attention is all you need.
\newblock In I. Guyon, U.~V. Luxburg, S. Bengio, H. Wallach, R. Fergus, S.
  Vishwanathan, and R. Garnett, editors, {\em Advances in Neural Information
  Processing Systems}, volume~30. Curran Associates, Inc., 2017.

\bibitem{misVeachPaper}
Eric Veach and Leonidas~J. Guibas.
\newblock Optimally combining sampling techniques for monte carlo rendering.
\newblock In {\em Proceedings of the 22nd Annual Conference on Computer
  Graphics and Interactive Techniques}, SIGGRAPH '95, page 419–428, New York,
  NY, USA, 1995. Association for Computing Machinery.

\bibitem{SSIMPaper}
Zhou Wang, Alan~C Bovik, Hamid~R Sheikh, and Eero~P Simoncelli.
\newblock Image quality assessment: from error visibility to structural
  similarity.
\newblock {\em IEEE transactions on image processing}, 13(4):600--612, 2004.

\bibitem{whitted2005improved}
Turner Whitted.
\newblock An improved illumination model for shaded display.
\newblock In {\em ACM Siggraph 2005 Courses}, pages 4--es. 2005.

\bibitem{sphericalHarmonicsRecent}
Lifan Wu, Guangyan Cai, Shuang Zhao, and Ravi Ramamoorthi.
\newblock Analytic spherical harmonic gradients for real-time rendering with
  many polygonal area lights.
\newblock {\em ACM Trans. Graph.}, 39(4), July 2020.

\bibitem{spaceTimeNeRF}
Wenqi Xian, Jia-Bin Huang, Johannes Kopf, and Changil Kim.
\newblock Space-time neural irradiance fields for free-viewpoint video.
\newblock {\em arXiv preprint arXiv:2011.12950}, 2020.

\bibitem{zhang2020nerf++}
Kai Zhang, Gernot Riegler, Noah Snavely, and Vladlen Koltun.
\newblock Nerf++: Analyzing and improving neural radiance fields.
\newblock {\em arXiv preprint arXiv:2010.07492}, 2020.

\bibitem{LPIPSPaper}
Richard Zhang, Phillip Isola, Alexei~A. Efros, Eli Shechtman, and Oliver Wang.
\newblock The unreasonable effectiveness of deep features as a perceptual
  metric.
\newblock {\em CoRR}, abs/1801.03924, 2018.

\bibitem{zhong2020reconstructing}
Ellen~D. Zhong, Tristan Bepler, Joseph~H. Davis, and Bonnie Berger.
\newblock Reconstructing continuous distributions of 3d protein structure from
  cryo-em images, 2020.

\end{thebibliography}
